\documentclass[12pt]{article}
\usepackage{setspace}
\usepackage{graphicx}
\usepackage{parselines} 
\usepackage{amsmath}
\usepackage{subfig}
\usepackage[left=25mm, right=25mm, top=25mm, bottom=20mm]{geometry}

\begin{document}  

\title{Towards Self-constructive Artificial Intelligence: Algorithmic basis (Part I)}  

\author{Fernando J. Corbacho
\\~\\~\\Cognodata I+D (R \& D)
\\\&
\\Computer Science Department. Universidad Aut\'onoma de Madrid.
\\~\\fernando.corbacho@cognodata.com}

\maketitle
\def\baselinestretch{1.7}

\singlespacing

\begin{abstract}
\def\baselinestretch{1.7}
Artificial Intelligence frameworks should allow for ever more autonomous and general systems in contrast to 
very narrow and restricted  (human pre-defined) domain
systems, in analogy to how the brain works. 
Self-constructive Artificial Intelligence ($SCAI$) is one such possible framework. 
We herein propose that $SCAI$ is based on three principles of organization: self-growing, self-experimental and self-repairing. 
Self-growing: the ability to autonomously and incrementally construct structures and functionality as needed to solve encountered (sub)problems.
Self-experimental: the ability to internally simulate, anticipate and take decisions based on these expectations. 
Self-repairing: the ability to autonomously re-construct a previously successful functionality or pattern of interaction lost from a possible sub-component failure (damage). 
To implement these principles of organization, a constructive architecture capable of evolving adaptive autonomous agents is required.
We present Schema-based learning as one such architecture capable of incrementally constructing a myriad of internal models of three kinds: predictive schemas, dual (inverse models) schemas and goal schemas as they are necessary to autonomously develop increasing functionality. 

We claim that artificial systems,
whether in the digital  or in the physical world,
can benefit very much form this  constructive architecture and should  be organized around these principles of organization. 
To illustrate the generality of the proposed framework,
we include several test cases in 
structural adaptive navigation in artificial intelligence systems  in Paper II of this series,
and resilient robot motor control in Paper III of this series.
Paper IV of this series will also include $SCAI$ for problem structural discovery in predictive Business Intelligence.

\end{abstract}

\def\baselinestretch{2}
\small

\begin{singlespacing}


\tableofcontents

\newpage

\section{Introduction}

As it is generally accepted, Artificial Intelligence (AI) is intelligence exhibited by machines, in contrast with the natural intelligence  displayed by humans and other animals (Wikipedia). In computer science, an ideal "intelligent" machine is a flexible rational agent that perceives its environment and takes actions that maximize its chance of success at some goal (Legg \& Hutter, 2007; Nilsson, 1998; Russell \& Norvig, 2010).
The central problems (or goals) of AI research include reasoning, knowledge, planning, learning, natural language processing (communication), perception and the ability to move and manipulate objects 
(Luger \& Stubblefield 2004; Nilsson 1998; Poole, Mackworth \& Goebel 1998; Russell \& Norvig 2010).
General intelligence is among the field's long-term goals (Kurzweil, 1999; 2005; Goertzel, 2014), thus, artificial Intelligence frameworks should allow for ever more autonomous and general systems in contrast to very narrow and restricted domain operating systems, in analogy to how the brain works. 
This paper presents Self-constructive Artificial Intelligence (SCAI) as one such possible framework for artificial general intelligence.

Integrative paradigms based on agents (Nilsson, 1998; Rusell \& Norvig, 2010) are becoming a central thrust of research in AI. 
In this regard, section 1.4 presents the agent-based paradigm. 
In adaptive agents, learning must be an integral aspect of all the AI components (Nilsson, 1998) and not an add-on component, as it is reflected in many frameworks for the design of AI agents (Albus, 2001; Bach, 2009; Goertzel et al., 2010; Hawkins, 2004; 2007; Kurzweil, 2012; Rosenbloom, 2013).
At the most advanced level, autonomous learning from the environment (Shen, 1994) becomes a central ingredient. 
There are nowadays several cases where autonomy is already playing a central role. A couple of representative examples include:
self-learning, referring to  ever more autonomous machine learning and self-driving, referring to autonomous vehicle navigation.

To illustrate the generality of the proposed framework,
we include several test cases in 
structural adaptive navigation in artificial intelligence systems,
and resilient robot motor control.
Part II of this series of papers will also include $SCAI$ for problem structural discovery in Business Intelligence.
In this paper we introduce three very different test cases to prove the generality of the SCAI framework.
Navigation in physical space is becoming paradigmatic (cf. self-driving in autonomous vehicles). 
We develop structural adaptive navigation to exemplify $SCAI$ in an incremental evolutionary AI. 
We have already  taken some of these aspects to Robotics (Weitzenfeld, Arkin, Cervantes-Perez, Olivares \& Corbacho, 1998). 
For more detail on some of the test cases we provide a more detailed description elsewhere (Corbacho \& Arbib, 2019; Sanchez-Montanes \& Corbacho, 2019). 
Next, we develop a case in resilient motor control where a previously reach and grip behavior is re-constructed after a break-up of the structure and functionality has taken place. 
In Part II of this series of papers, we also present structural Learning for autonomous problem structural decomposition as an ingredient of SCAI. 
Real world problems typically present a complex structure. Much of the work of the researcher consists of uncovering this underlying structure which allows to decompose the problem into easier to cope with and solve sub-problems.
Subsequently, each specific subproblem has a particular scope and boundary conditions.

\subsection{Artificial General Intelligence, Developmental and Evolutionary Artificial Intelligence}

As already expressed, general intelligence is among the field's long-term goals (Kurzweil, 1999; 2005). In this regard, 
SCAI is a framework for general intelligence  in contrast to very narrow and restricted domains of application. 
Artificial Intelligence frameworks should allow for ever more autonomous and general systems in contrast to very narrow and restricted domain operating systems, in analogy to how the brain works. 
Self-constructive Artificial Intelligence (SCAI) is one such possible framework. 

As Goertzel  and Pennachin (2007) expressed, in recent years a broad community of researchers has emerged, focusing on the original ambitious goals of the AI field - the creation and study of software or hardware systems with general intelligence comparable to, and ultimately perhaps greater than, that of human beings. 
Goertzel (2010; 2014) review approaches to defining the concept of Artificial General Intelligence (AGI) including mathematical formalisms, engineering, and biology inspired perspectives. 
The spectrum of designs for AGI systems includes systems with symbolic, emergentist, hybrid and universalist characteristics. 
The question is how to best conceptualize and approach the original problem regarding which the AI field was
founded, 
that is, the creation of thinking machines with general intelligence comparable to, or greater than, that of human beings.
The standard approach of the AI discipline (Russell and Norvig, 2010), as it
has evolved in the six decades since the field’s founding, views artificial intelligence largely in terms
of the pursuit of discrete capabilities or specific practical tasks. 
But while this approach has yielded
many interesting technologies and theoretical results, it has proved relatively unsuccessful in terms
of the original central goals of the field.

In this regard, Ray Kurzweil (Kurzweil, 2005) has used the term “narrow AI” to refer to the creation of
systems that carry out specific “intelligent” behaviors in specific contexts. For a narrow AI system,
if one changes the context or the behavior specification even a little bit, some level of human
reprogramming or reconfiguration is generally necessary to enable the system to retain its level
of intelligence. This is quite different from natural generally
 intelligent systems like humans, which
have a broad capability to self-adapt to changes in their goals or circumstances, performing “transfer
learning” (Taylor, Kuhlmann, and Stone, 2008) to generalize knowledge from one goal or context to
others. The concept of “Artificial General Intelligence“ has emerged as an antonym to “narrow AI”,
to refer to systems with this sort of broad generalization capability. 
The AGI approach takes
“general intelligence“ as a fundamentally distinct property from task or problem specific capabilities,
and focuses directly on understanding this property and creating systems that display it. 
One of the long  term goals of general intelligence is achieving commonsense knowledge. 
We claim that commonsense knowledge is reached by self-experienced/self-experimented knowledge and is grounded by 
sensorimotor interactions with the world, much as a ten-years old child has developed its commonsense knowledge of the world (Nilsson, 1998). 
For this SCAI at the subsymbolic level is needed to be able to ground concepts in the real world represented by signals. 
We will later show this in several examples on navigation and motor control implemented by schemas operating over signals.

SCAI is also developmental intelligence. A clear early example of developmental AI is  developed by Drescher (1991). 
He described a theory of how a computer program might be implemented to learn and use new concepts that have not been previously programmed into it.
Drescher described the schema mechanism, a general learning and concept-building mechanism (at the symbolic level) inspired by Jean Piaget's account of human cognitive development (Piaget, 1954), 
The schema mechanism is intended to replicate key aspects of cognitive development during infancy. It takes Piaget's theory of human development as source of inspiration for an artificial learning mechanism; and it extends and tests Piaget's theory by checking whether a specific mechanism that works according to Piagetian themes actually exhibits Piagetian abilitiehe schema mechanism learns from its experiences, expressing discoveries in its existing representational vocabulary, and extending that vocabulary with new concepts. A novel empirical learning technique, marginal attribution, can find results of an action that are obscure because each occurs rarely in general, although reliably under certain conditions. Drescher shows that several early milestones in the Piagetian infant's invention of the concept of persistent object can be replicated by the schema mechanism.

A related field of study, developmental robotics, sometimes called epigenetic robotics, is a scientific field which aims at studying the developmental mechanisms, architectures and constraints that allow lifelong and open-ended learning of new skills and new knowledge in embodied machines. As in human children, learning is expected to be cumulative and of progressively increasing complexity, and to result from self-exploration of the world in combination with social interaction.
Internal Models (described in later sections)  and dynamic representations (Weng et al., 2001) play a significant role in autonomous developmental robotics. 
They are mechanisms able to represent the input-output characteristics of the sensorimotor loop. 
In developmental robotics, open-ended learning of skills and knowledge serves the purpose
of reaction to unexpected inputs, to explore the environment and to acquire new
behaviors. The development of the robot includes self-exploration of the state-action
space and learning of the environmental dynamics (Smith-Bize, 2016)

SCAI is also evolutionary AI (Nilsson, 1998).
In this regard, developmental robotics (DR)  is related to, but differs from, evolutionary robotics (ER). ER uses populations of robots that evolve over time, whereas DR is interested in how the organization of a single robot's control system develops through experience, over time.
In this regard work from Rana computatrix (Arbib, 1987; Corbacho \& Arbib, 1996) to Rattus computator (Guazzelli, Corbacho, Bota \& Arbib, 1998) provides an example of evolutionary AI in which certain principles of sensorimotor integration are taken and evolved from models of amphibian  behavior to the world of mammals (concretely to rodents). 
This paper in the series presents SCAI principles of organization that can be already found in navigation and motor control present in amphibian and, hence, are also present and are the basis for more complex behaviors in mammals (Paper III in this series).

\subsection{Self-constructive Brain}

We have already proposed that the emergence of adaptive behavior in animals arises from the self-constructive nature of the brain (Corbacho, 2016). 
The brain builds itself up by reflecting on its particular interactions with the environment, 
that is, it constructs its own interpretation of reality through the construction of 
representations of relevant predictive patterns of sensorimotor interaction. 
Thus,  we presented the brain as the architect of its own reality, that is, the brains builds reality rather than simply observing it. 
We introduce the predictive (forward internal models) and its associated dual schemas as active processes capturing relevant patterns of sensorimotor interaction; and we suggest that the brain is composed of  myriads of these patterns.
These predictive internal models exist all over the brain and a variety of examples can be found in the literature regarding different brain areas/functionalities (cf. literature on the predictive brain: Clark, 2013; Corbacho, 2016).

We have proposed that the brain is self-constructive (SCB) since it is self-experimental, self-growing, and self-repairing.
The brain is self-experimental since to ensure survival the self-constructive brain is an active machine capable of performing experiments of its own interactions with the environment 
as well as capable of mentally simulating the results of those interactions in order to be able to later decide the most optimal course of action. 
In this regard, the way for our brain to fully understand anything is to model and simulate it. 
To survive it must anticipate since anticipating an event allows to better prepare for it, since it allows the animal/agent to get ready to act immediately with the most successful course of action possible.
Anticipation  plays an important role in directing intelligent behavior under a fundamental hypothesis that 
the brain constructs reality as much as it embodies it. 
Hence, the anticipatory nature of the brain is a clear ingredient necessary for the SCB.
The brain is also self-growing, since it dynamically and incrementally constructs internal structures in order to build a model of the world as it gathers statistics.
To do so, it incrementally recruits and then adapts different neural circuitry that implements different behavioral modules (schemas). 
Finally, the brain is also self-repairing, since to survive, it must also be robust and capable of self-organization and self-repair, that is, the brain is capable of finding ways to repair parts of previously working structures and hence re-construct a previous successful pattern of interaction.  This is the case when brain lesions occur.

These are evolutionary principles that have evolved in many different species, not just in mammals. That is, these principles can be clearly observed at different levels of abstraction and simple implementations can be  found even in lower vertebrates. 
So, for instance, amphibian brain capacity for recovering a behavioral pattern of interaction after a critical lesion demonstrates their capacity for self-repair. Self-experimental and self-growing, on the other hand, can also be tested by their ability to learn certain problems such as learning to detour through the use of internal models.

\subsection{Agent-based Artificial Intelligence}

Nilsson (1998) provides a synthetic view of AI based on agents. 
In this regard, artificial intelligence agents are composed of combinations of the following components: 
Perception, integration of perception and action, control of action, planning and learning.
In a similar fashion, Russel \& Norvig (2010) main unifying theme is the idea of an intelligent agent. 
They define AI as the study of
agents that receive percepts from the environment and perform actions. Each such agent implements
a function that maps percept sequences to actions. 
They explain the role of learning as extending the reach of the designer into unknown
environments, They also treat robotics (action) and vision (perception) not as independently defined
problems, but as occurring in the service of achieving goals. Also they stress the importance of the
task environment in determining the appropriate agent design.

In this view, an agent is anything that can be viewed as perceiving its environment through sensors and
acting upon that environment through actuators. 
Russel \& Norvig (2010) use the term percept to refer to the agent’s perceptual inputs at any given instant. 
An agent's percept sequence is the complete history of everything the agent has ever perceived.
In general, an agent's choice of action at any given instant can depend on the entire percept
sequence observed to date, but not on anything it has not perceived. By specifying the agent's
choice of action for every possible percept sequence, we have said more or less everything
there is to say about the agent. Mathematically speaking, we say that an agent's behavior is
described by the agent function that maps any given percept sequence to an action.
Internally, the agent function for an artificial agent will be implemented by an
agent program. It is important to keep these two ideas distinct. The agent function is an
abstract mathematical description; the agent program is a concrete implementation, running
within some physical system.

They answer this age-old question in an age-old way: by considering the consequences
of the agent's behavior. When an agent is plunked down in an environment, it generates a
sequence of actions according to the percepts it receives. This sequence of actions causes the
environment to go through a sequence of states. If the sequence is desirable, then the agent
has performed well. This notion of desirability is captured by a performance measure that
evaluates any given sequence of environment states.
As a general rule, it is better to design performance measures according to what one actually
wants in the environment (cf. goal states), rather than according to how one thinks the agent should behave
(Russel \& Norvig, 2010). 
To the extent that an agent relies on the prior knowledge of its designer rather than
 on its own percepts, we say that the agent lacks autonomy. A rational agent should be
autonomous: it should learn what it can to compensate for partial or incorrect prior knowledge.
Russel \& Norvig (2010)  group all these under the heading of the task environment. 
They call this the PEAS (Performance, Environment, Actuators, Sensors) description. In designing an
agent, the first step must always be to specify the task environment as fully as possible.
Section 5.1 in paper II of this series will provide  descriptions of some of the environments and their dynamics.

\subsubsection{Agent formalization and dynamics}

Arbib (1987) already argued that to understand complex behavior it will not be satisfactory to use the classical description of finite automata that specifies the next state for every state-input pair. Rather, we must see how various small automata are built up one upon the other in a hierarchical fashion. 
We now formally define an adaptive autonomous agent as a hierarchical network of port automatas (Corbacho, 1997). 
We have tried to stay away from the "global" state-space representation employed to formalize autonomous agents (Rivest and Shapire, 1989; Shen, 1994) to a more "diversity-based" modular representation closer to the structural development and composition of biological systems. Thus, an adaptive autonomous agent ($A$) is an automata network (Lyons \& Arbib, 1989; Steenstrup, Arbib \& Manes, 1983; Weisbuch, 1991) composed of a set of automata interconnected such that the outputs of some are the inputs to others. 
We will show that the hierarchical network of automata is a hierarchical network of different types of schemas (Arbib, 1995; Corbacho \& Arbib, 1996). 
That is,  $A$ inner's dynamics is also defined by multi-agent-based modelling, that is a network of simple interacting components gives rise to the emergence of different (complex) behaviors by competitive and cooperative distributed dynamics. 

To illustrate the main components we will use a particular formal implementation based on schema theory. 
Schema theory can be implemented by neural networks or can be implemented by any other language when applied to software agents or mechanical devices. Schema-based learning (SBL) is an extension to schema theory dealing with the autonomous construction of new schemas as they are needed to solve encountered problems. 
SBL is an architecture to construct a myriad of internal models of several types: predictive schemas, dual (inverse models) schemas and goal schemas on the run.
\\
\\
{\bf Definition. An Adaptive Autonomous Agent}  ($A$) is defined at a particular point in time $t$ by the tuple:
\begin{equation}
A(t) = (D,  S(t), C(t), Q(t), R(t), L)
\end{equation}
where  $D$ defines the different internal drives,  $D = \{d_i, ...\}$; $d_i = \{\alpha_i, d_i^{max}, a(i), I(i)\}$,
where $I(i)$ correspond to incentives and $a(i)$ correspond to drive reduction in the presence of some substrate. 
$S(t) = \{S^1, ..., S^N\}$ determines the set of shemas $S^1, ... , S^N$ at time $t$;
$C(t)$ determines the  port-to-port connection map; $Q(t)$ determines the  schema support network; $R(t)$ determines the cause-effect relations between different schemas, typically initially empty since they are learned by observing the effects of the interactions after they have been experimented by the agent, and $L$ corresponds to the constructive algorithm in charge of constructing new schemas in the set $S$ as well as their new relations in $C, Q, R$.

\begin{figure}[ht!]
\centering
\includegraphics[width=120mm]{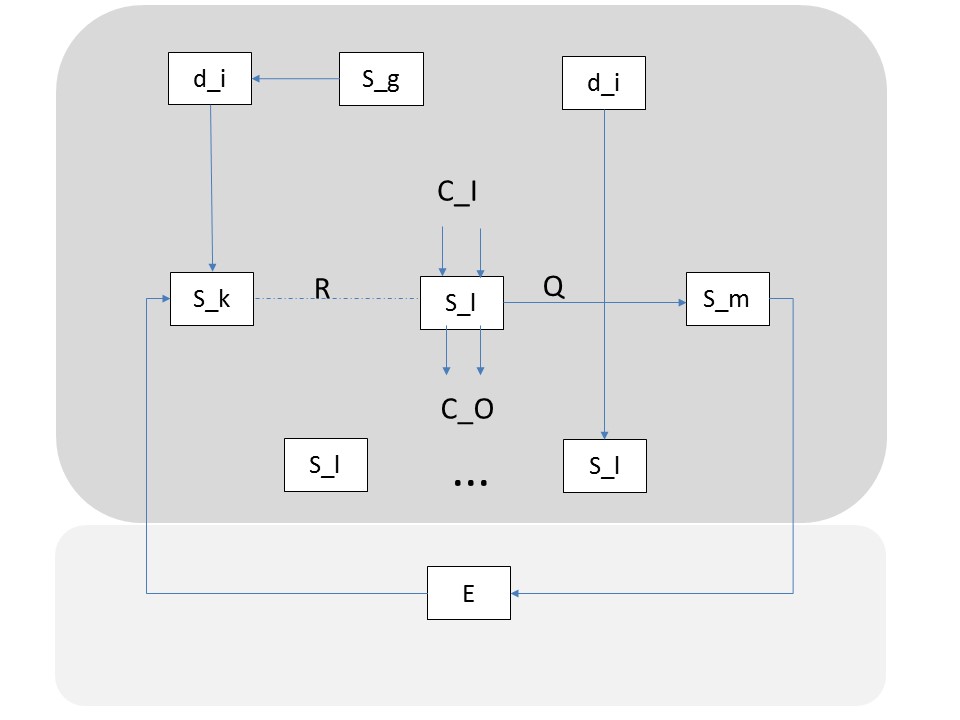}
  \caption{{\bf The agent immersed in the environment E.}
The agent receives input from the environment through the receptors in primitive perceptual schemas (e.g. $S_k$) and affects the environment through the effectors in primitive motor schemas (e.g. $S_m$). All sorts of internal interactions take place defining in some sense another "internal environment". The different internal drives affect the inner dynamics (e. g. by setting goals to be achieved). This diagram only illustrates one example for each of the different possible internal interactions}
  \label{figAgentinEnvironment}
\end{figure}

A particular Agent is related to a particular type of environment. 
The environment $E$ directly interacts with the autonomous agent $A$ through the agent's receptors which receive signals from $E$ and 
the agent´s effectors which, in turn, affect $E$ (Figure \ref{figAgentinEnvironment}).
Both dynamics are captured by the primitive perceptual  and primitive motor schemas respectively (both in the set $S(t)$).
\\
- Primitive perceptual schemas:  the input port of schema $S^k$ : ($input_i^k = f(E(o_j)))$
\\
- Primitive motor schemas schemas:  the output port of schema $S^m$ : $( E(i_l) = g(output_i^m))$

Thus, there is a clear analogy with respect to the PEAS framework of Russel \& Norvig (2010) described in section 1.3. In this regard, the primitive perceptual schemas are in direct connection to the agent's sensors (S in PEAS). The primitive action schemas are in direct connection to the actuators (A in PEAS). 
The environment function E is modeled in SCAI  by yet another set of port automata.
Finally, the performance function (P) is formalized in SCAI  by drive reduction, value maximization, performance and coherence maximization. 
SCAI will incrementally extend and adapt $S(t)$, $C(t)$, $Q(t)$, and $R(t)$ accordingly in order to 
maximize coherence with its environment. 

We will later show how the network of interacting schema instances will take the AAA from a situation (context), and goals (desirable states triggered by drives) to a feasible course of action that meets the goals. The set of schemas changes through time, thus, the Connectivity map has to change appropriately. Now let us proceed to formalize schemas within the schema-based learning constructive architecture.

\section{Self-constructive Principles of Organization}
We propose that SCAI is based on three principles of organization: self-growing, self-experimental and self-repairing; 
and these principles apply to all aspects of intelligence: 
perception, integration of perception and action, control of action, and planning. 
Learning already being an integral part of the previous components. 
We also suggest that AGI must also, at least, follow these same three principles of organization.

Self-growing: the ability to autonomously and incrementally construct structures and functionality as needed to solve encountered (sub)problems.
Self-experimental: the ability to internally simulate, anticipate and take decisions based on these expectations. 
Self-repairing: the ability to autonomously re-construct a previously successful functionality or pattern of interaction.

Self-experimental: an active machine capable of performing experiments of its own interactions with the environment 
as well as capable of internally simulating the results of those interactions in order to be able to later decide the most optimal course of action. 
In order to understand its relation with the environment, the agent constructs internal models of reality that allow the agent to internally simulate its interactions with the environment. 
Hence, the agent has the capacity to design experiments to be performed by  interacting with the environment as well as  experiments inside the mind, for instance by mental practice/modeling, visualization and simulation. 
In this regard, cause-effect experimentation (section 4.1) triggers the construction of predictive schemas and its dual  schemas.  Hence, a very important component of the self-experimental agent consists on the ability to simulate the causal flow of the interactions with the environment. 
In this regard, predictive schemas allow for internal modeling. 
That is, they allow the system to anticipate the results of an action before it is taken. That is, predictive schemas produce anticipatory representations of events/effects that have no yet occurred in the external environment. 

The self-constructive agent is self-growing. In this regard we show an example of schema construction growing a new topological configuration to represent, and hence be able to reproduce, a successful relevant pattern of interaction. 
We show how the agent builds itself up using the SBL formalization of schema construction. 
This relates to structural learning (Kemp \& Tenembaum, 2008; Sanchez-Montanes \& Corbacho, 2019; Salakhutdinov, Tenenbaum, \& Torralba, 2013; Tenembaum, Kemp, Griffiths \& Goodman, 2011) and 
modular learning (Jacobs, Jordan, Nowlan \& Hinton, 1991; Jordan \& Jacobs, 1992) since $SCAI$ autonomously constructs new modular structures that allow increasing functionality while encapsulating existing functionality, and avoiding destructive interference across different functional tasks. 

We finish up by showing that the self-constructive agent is self-repairing.
Thar is, the agent has the capacity to repair itself and re-construct past functionality after certain functionality may have been damaged. 
In order to so, we show how the constructive principles developed are able to explain learning after damage of the control circuitry that initially prevents grip closing during reach and grasp motor functionality (critical for many robots).

\subsection{Self-constructive Artificial Intelligence  is self-experimental}
Intelligent agents are self-experimental since to ensure "survival" the self-constructive agent is an active machine capable of performing experiments of its own interactions with the environment 
as well as capable of mentally simulating the results of those interactions in order to be able to later decide the most optimal course of action. 
As previously described, different schema molecules are activated in parallel to select the proper course of action. 
The right dual schema is selected whose predictive schema simulates and anticipates the desired goal state. 

In this regard, the way for our mind to fully understand anything is to model and simulate it. 
As Pezzulo (2008) has expressed, cognition is for doing, for simulating. 
In order to understand its relation with the environment the mind constructs internal models of reality that allow the mind to internally simulate its interactions with the environment. 
Hence, the intelligent agent has the capacity to design experiments to be performed by  interacting with the environment as well as  experiments inside the mind, for instance by mental practice, visualization and simulation. 
There is a growing body of experimental data that supports the idea of the self-experimental brain. 
In this regard,  experimental evidence indicates that animals can use mental simulation to make decisions about the action to take during goal-directed navigation (Chersi, Donnarumma, \& Pezzulo, 2013). Its most salient characteristic is that choices about actions are made by  simulating movements and their sensory effects using the same brain areas that are active during overt execution. 
Chersi et al. (2013) link these results with a general framework that sees the mind as a predictive device that can “detach” itself from the here-and-now of current perception using mechanisms such as episodic memories, motor and visual imagery. 
In this regard, the concept of action simulation is gaining momentum in cognitive science, neuroscience, and robotics, and in particular within the study of grounded, embodied and motor cognition (Declerck, 2013;  Hesslow, 2012; Raos, Evangeliou \& Saraki, 2007; Jeannerod, 2001; Mohan, Sandini \& Morasso, 2014; 
Pezzulo, Candini, Dindo, \& Barca, 2013). 

The ability to construct a hypothetical situation in one's imagination prior to it actually occurring may afford greater accuracy in predicting its eventual outcome. The recollection of past experiences is also considered to be a re-constructive process with memories recreated from their component parts (Hassabis \& Maguire, 2009). Construction, therefore, plays a critical role in allowing us to plan for the future and remember the past. Conceptually, construction can be broken down into a number of constituent processes although little is known about their neural correlates. Moreover, it has been suggested that some of these processes may be shared by a number of other cognitive functions including spatial navigation and imagination. 

In this regard, mental practice is the cognitive rehearsal of a physical skill in the absence of overt physical movement (Jordan, 1983). 
The questions arises whether mental practice enhances performance (Driskell, Copper \& Moran, 1994; Gentili, Han, Schweighofer \& Papaxanthis, 2010; Miall \& Wolpert, 1996).  
Mental practice promotes motor anticipation as for example there is evidence from skilled music performance.
On the other hand, mental imagery can be described as the maintenance and manipulation of perception and actions of a covert sort, i. e., it arises not as a consequence of environmental interaction but is created internally by the mind (Di Nuovo, Marocco, Di Nuovo \& Cangelosi; 2013; Kossyln, Gani \& Thompson, 2001; Lallee \& Dominey, 2013; Svensson, Thill \& Ziemke, 2013).
According to the simulation hypothesis, mental imagery can be explained in terms of predictive chains of simulated perceptions and actions, i.e., perceptions and actions are reactivated internally by our nervous system to be used in mental imagery and other cognitive phenomena (Chersi, Donnarumman \& Pezzulo, 2013; Hesslow, 2002, 2012; Svensson, Thill \& Ziemke, 2013). In this regard, Svensson, Thill \& Ziemke (2013) go a step further and  hypothesize that dreams informing the construction of simulations lead to faster development of good simulations during waking behavior.

Certain type of generative models such as the wake-sleep type algorithms (Hinton, Dayan, Frey \& Neal, 1995) are composed of forward connections as well as backward connections which can give rise to mental visualizations (Friston, 2010). 
That is, during the sleep generative phase, phantasy patterns corresponding to visualizations can be generated thanks to the information flow generated by the backward projection. 
In the same way, we show that predictive schemas are capable of producing anticipatory activity patterns as the result of mental modeling. 
In this regard, we will later show anticipatory (phantasy) activity pattern in the motor heading map ($\hat{mhm}(t+1)$) during learning to detour (cf. dynamic remapping: Droulesz \& Berthoz, 1991; Duhamel et al., 1992)
Also later we will display  internally simulated activity patterns in proprioceptive grip receptor schemas 
($\hat{grip\_rec}(t+1)$) 
during learning after a system failure has taken place. 
The prediction of states of the body has also been shown to be a useful capability in resilient robots (Bongard, Zykov, \& Lipson, 2006) and it has been suggested that this could be generalized to internal models of the environment (Adami, 2006). 

We have already introduced internal models in section 3.1 and their SBL implementation, predictive and dual schemas in section 3.2. Predictive schemas allow for mental modeling. 
That is, they allow the system to anticipate the results of an action before it is taken. That is, predictive schemas produce anticipatory representations of events/effects that have no yet occurred in the outside environment. 
That is, the self-experimental brain produces anticipatory experimental patterns of activity.
Yet the question remains as to how the predictive schemas are constructed. 
In this regard, cause-effect experimentation triggers the construction of predictive schemas and its corresponding dual schemas.  
Hence, a very important component of the self-experimental brain consists on the ability to simulate the causal flow of the interactions with the environment and, thus, learn cause-effect relations from the potentially combinatorially large set of hypothesis. 
In learning new causal relations by experimentation, the potentially combinatorial explosion is constrained by interacting with the environment. In an analogous manner as constraint relaxation dynamics reduce an exponential space into a more tractable space of possible configurations, section 4.1   will introduce the mechanisms to learn the effective 
cause-effect relations within a complex system (e.g. brain, autonomous agent).

\section{Self-constructive Architecture }

Schema theory and its extension with Schema-based learning (SBL from now on) may serve as a possible framework to formalize SCAI. 
Historically the notion of "internal models of the world" led to the notion of schema (Arbib, 1972; Barlett, 1932; Neisser, 1976, Piaget, 1971). 
We first introduce and formalize the notion of schema to describe the modular functional and structural characterization of the agent's mind.
Schemas also serve to encapsulate specific patterns of interaction (cf. Hinton et al., 2011; Sabour et al., 2017)
Then, we introduce the notion of predictive schema to generalize and formalize the concept of internal forward models within the SBL architecture (Corbacho and Arbib 1997c; 
Corbacho, 1998).
The work presented here emphasizes the use and construction of 
new predictive internal models (predictive schemas) as well as 
the construction of their corresponding dual schemas when  specific conditions arise during the 
interaction of the agent with its environment. 
SBL also presents a more general 
approach including a wider variety of schemas, 
and a larger repertoire of processes to construct the different kinds of schemas
under various conditions. 

Corbacho and Arbib (1997a) introduced the notion of 
{\bf coherence}
to emphasize the importance of maximizing the  congruence between 
the results of an interaction (external or internal) and the expectations
(previously learned) for that interaction. 
The fundamental two principles of organization  in SBL are coherence and performance maximization.
Besides the main units of organization, the predictive schema and its dual associated schema, goal schemas are the other component in charge of dynamically setting a hierarchy of goals. 
SBL attempts to reduce incoherences and get closer to goal states simultaneously.
One of the main operations in SBL is the construction of all these internal models. 

We will briefly review the
notion of schema introduced in Corbacho (1997). 
We  described a schema  as a
unit of concurrent processing corresponding to a domain of interaction.
Lyons \& Arbib (1989) provided a formal semantics based on port
automata and Corbacho (1997) extended this definition to include 
schema activity variables and their dynamics. Other schema formalizations related to development and learning have been proposed (e.g. Drescher, 1991; Pezzulo, et al., 2013)). 
\\
\\
{\bf Definition:} A basic {\it schema} description is 
\begin{verbatim}
  basic-schema::= [Schema-Name:   <N>
  Input-Port-List:               (<Iplist>)
  Output-Port-List:              (<Oplist>)
  Variable-List:                 (<Varlist>)
  Behavior:                      (<Behavior>)]
\end{verbatim}	

where
$N$ is an identifying name for the schema $S^N$,

$<Iplist>$ and $<Oplist>$ are lists of $(Portname: Porttype)$ pairs
for input and output ports, respectively.

$<Varlist>$ is a list of $(Varname:Vartype)$ pairs for all internal
variable names, and 

$<Behavior>$ is a specification of computing behavior. 
\\
\\
The notation $i^{x}_k(t)$ is used to represent the
pattern of activity in the $k$th
input port of schema $S^{x}$ at time $t$ and
  $o^{x}_k(t)$ for the analogous output port.
In the rest of the paper, in order to alleviate the use of notation,
when the schema has a single output port, then this output port takes the name of the
schema in lower case. That is, for instance we will use the notation $target(t)$  to
name the output port activity pattern from the $S^{TARGET}$ target recognition visual schema.

As mentioned before rather than simply process input symbols to yield output symbols, the individual schemas have activation levels which measure some degree of confidence, and it is, for instance, the more active of a set of perceptual schemas that will trigger the appropriate motor schema(s) to yield the appropriate response. This is a very simple example of the type of mechanisms of competition and cooperation (Arbib, 1995) that can be exhibited by a network of schemas. In particular multiple motor schemas may be coactivated to control subtle behaviors. 
Thus, we have extended the definition of schema to allow for activation variables. We will also introduce an activation rule defining the dynamics of the activation variable. Other schemas may affect the state of the activation variable (sending support through their respective control output ports), thus, the need for both input and output control ports. These ports are called control ports since they determine the state of the activation variable which in turn determines whether to "activate" the (standard) output ports. In summary theActivation Rule  updates the activation variable based on support from other schemas. Also each schema has an associated threshold    $\theta^x \in \Re$ to determine when the schema  is activated. Consider the schema $S^x$: 
 \\

{\bf AcVarlist} is a list of pairs containing $(a_m^x: \Re), (q^{x,y}: \Re), (\theta^x: \Re)$.
\\

{\bf CIolist} is a lists  of $(Portname: \Re)$ pairs for input control ports.
\\

{\bf COolist} is a pair $(Portname: \Re)$  for output control ports.
\\

{\bf AcRule} is a specification of computing behavior
\begin{equation}
a^x_m(t+1) = \sigma(\sum_y q_{x,y}(t)  \cdot a^y_m(t))
\end{equation} 
where  $q_{x,y} \in \Re$ "weights" the support from schema $S^y$  to schema$S^x$.  

These new components may be introduced as farther "slots" in the RS basic schema structure or they can be incorporated within the already existing "slots". The second choice might already provide hints as to the semantic mapping from this extended schema to the Port Automaton (PA). Namely, the activation variable can be considered an internal variable of the schema in the RS terminology. The control ports may be added to the list of ports, and the Activation Rule becomes extra statements in the Behavioral specification.
\\
\\
Schema assertion is defined in RS by default, that is, when a schema was instantiated in RS it was immediately asserted. Hence, here we introduce another state which corresponds to the case when a schema that is instantiated is deasserted (not asserted).
\\
\\
{\bf Definition}. Schema Deassertion. The activation rule  decides when to deactivate output ports. An SI of the schema $S^x$,  $S^x_v(i_1, i_2, ...)(o_1, o_2, ....)$,   is deasserted iff $a^x_n \le 0$, then for all $i, o_i = \sharp$. Following Lyons and Arbib's (1989) convention, a value of $\sharp$ indicates that there is no input or output at the designated port. 
As a corollary we can see that when a schema is not asserted it can not support other schemas since its output control port will also be deactivated.
\\
\\
As we have already mentioned, Corbacho \& Arbib (1997c) also
introduced two special kinds of schemas, namely {\it predictive
schemas} and {\it goal schemas}. 
Goal schemas are a special kind
of schemas whose output port corresponds to a goal state (i.e. desired
pattern of activation in another schema). 
On the other hand, the role of the {\it predictive schema} is to anticipate the effect that the
particular pattern of activation of the {\it cause} schema has on the
current state of the {\it effect} schema.

\subsection{Internal models for predictive adaptive control}
The importance of internal models in intelligent behavior has been acknowledged for many years
(e.g. Craik 1943; Gregory, 1967; Arbib 1972).
Internal models predict the evolution of the environment by imitating its causal flaw.
They play an important role in directing intelligent behavior under a fundamental hypothesis that 
the mind constructs reality as much as it embodies it. 
At the behavioral level we see that animals learn to anticipate predictable events. 
The term internal model is also popular in control theory
and denotes a set of equations that describes the temporal development
of a real world process (e.g. Kalman 1960; Garcia et al. 1989).

The concept of an internal model, a system which mimics the behaviour of a natural process, has also emerged as an important theoretical concept in motor control (Haruno, Wolpert \& Kawato, 2001; Jordan, 1983; Kawato, 1990, 1999; Kawato, Furukawa \& Suzuki, 1987;  Miall \& Wolpert, 1996;  Wolpert, Ghahramani \& Jordan, 1995; 
Wolpert \& Kawato, 1998). 
Internal models can be classified into several conceptually distinct classes. 
One type of internal model is a causal representation of the motor apparatus, sometimes known as forward model (Jordan \& Rumelhart, 1992). Such a model would aim to mimic or represent the normal behavior of the motor system in response to outgoing motor commands. 
A forward model is a key ingredient in a system that uses motor outflow (also called efference copy) to anticipate and cancel the sensory effects of movement. The internal sensory signal needed to cancel reafference has been labeled corollary discharge. 

Potential uses of forward models include: canceling sensory reafference, distal supervised learning, internal feedback to overcome time delays, state estimation, and state prediction for model predictive control and mental practice/planning (Miall \& Wolpert, 1996). 
More specifically, predictive internal models may allow to
transform errors between the desired and actual sensory outcome for
a movement into the corresponding error in motor command,
to resolve ambiguous information,
to synthesize information from disparate sources,
to combine efferent and afferent information,
to perform mental practice (internal simulation) to learn to select
between possible actions,
to perform  state estimation in order to anticipate and cancel sensory
effects of ones own actions (not to distract attention/resources),
to reduce the credit assignment space,
to contribute to reasoning and planning by forming predictive chains, 
and finally to estimate the outcome of an action and
use it before sensory feedback is available, 
when delays make feedback control too slow for rapid movements 
(Miall \& Wolpert, 1996). 

One fundamental problem which the system  faces in the context of motor control is that the goal and outcome of a movement are often defined in task-related coordinates (Jordan \& Rumelhart, 1992). 
A basic problem, therefore, exists in translating these task-related (visual or auditory) goals and errors into the appropriate intrinsic signals (motor commands and motor errors) which are required to update the controller. The forward model can be used to estimate the motor errors during performance by backpropagation of sensory errors through the model. 
In this paper we will show how this problem of distal supervised learning generalizes beyond motor control and applies to many schemas all over the agent architecture.

Another kind of internal models are known as inverse models (Atkeson, 1989), which invert the causal flow of the motor system. They generate, from inputs about its state and state transitions, an output representing the causal events that produced that state (Cruse \& Steinkuelher, 1993; Wada \& Kawato, 1993; Shadmehr \& Mussa-Ivaldi, 1994).
For example, an inverse dynamics model of the arm would estimate the motor command that caused a particular movement. The input might therefore be the current and the desired state of the arm; the output would be the motor command which would cause the arm to shift from the current state to the desired state. An inverse sensory output model would predict the changes in state that corresponded to a change in sensory inflow. 
In the kinematic domain the inverse kinematic model again inverts the forward kinematic model to produce a set of joint angles which achieve a particular hand position. However, as a forward model may have a many-to-one mapping, there is no guarantee that a unique inverse will exist. 

As already mentioned, historically, the notion of "internal models of the world" led
to the notion of schema (Arbib, 1972). 
Hence, we will use schema theory and particularly its extension schema-based learning (Corbacho \& Arbib, 1997c) as the framework to formalize SCAI.

\subsection{Predictive and Dual schemas}
Corbacho (1997) presented predictive schemas as a generalization of
forward models (Jordan \& Rumelhart, 1992) since they not only apply to motor control but rather to perceptual, sensorimotor and abstract representational spaces as well. 
As already described, the role of the {\it predictive schema} is to anticipate the effect that the
particular pattern of activation of the {\it cause} schema has on the
current state of the {\it effect} schema. 
Also every predictive schema has
an associated {\it dual schema} that is responsible for selecting the optimal pattern of activity in the cause schema such that a goal pattern of activity in the effect schema can be achieved. 
\\
\\
{\bf Definition:} A {\it predictive schema} $P^{x,y}$ associated with 
effect schema $S^x$ and dual schema  $S^{y,x}$ (cause schema $S^y$) is a schema with the
following special characteristics:

$<Iplist>$: The first input port is of the same type and connects to the
output port of the effect schema $S^x$, i.e. $i^{x,y}_1(t)=o^{x}(t)$.
The second input port is of the same type and connects to the
output port of the dual schema $S^{y,x}$,
(cause schema $S^y$),
i.e. $i^{x,y}_2(t)=o^{y,x
}(t)$ ($i^{x,y}_2(t)=o^{y}(t)$).
The remaining input ports are optional and correspond to {\it context}
 information; i. e. $i^{x,y}_3(t)=o^{v}(t)$

$<Oplist>$: The output port contains the predictive response
$\hat{o}^{x,y}(t+1)$, representing the expectation for the state of the output port of the effect schema $S^x$ at time $t+1$, given the information in the input ports, that is, 
$\hat{o}^{x,y}(t+1) = E[o^{x}(t+1) | o^{x}(t), o^{y,x}(t), o^{v}(t)]$.  		

$<Varlist>$: Includes the parameters of the predictive schema, namely ${\bf W}^{x,y}_P (t)$.	

$<Behavior>$: The predictive schema behavioral specification includes a mapping $M^{x,y}_P$ parameterized by ${\bf W}^{x,y}_{P} (t)$, such that 
\begin{equation} 	
\hat{o}^{x,y}(t+1) = M^{x,y}_P (o^{x}(t), o^{y,x}(t), o^{v}(t); 
{\bf W}^{x,y}_{P} (t))
\label{eq_pred_sch}
\end{equation} 
 as well as a mapping $T_P$ that allows the predictive schema to be tuned, i.e. its parameters 
change according to the prediction error $(o^{x}(t+1) - \hat{o}^{x,y}(t+1))$,
so that the predictive response
$\hat{o}^{x,y}(t+1)$
becomes increasingly closer to the observed response 
$o^{x}(t+1)$
as the number of interactions increases. 
In this regard, several error minimization methods
are valid. Also, depending on the architectural implementation of the schema, 
${\bf W}^{x,y}_P$ takes the corresponding form.
In this article we use the mean squared error as the cost functional for schema adaptation as suggested by Jordan \& Rumelhart (1992). 
Hence, the functional becomes
\begin{equation}
{\bf W}^{x,y}_P (t+1)=T_P ({\bf W}^{x,y}_P (t), 
\hat{o}^{x,y}(t+1), 
o^{x}(t+1), 
\frac{\partial \hat{o}^{x,y}}{\partial {\bf W}^{x,y}_P })
\label{eq_pred_sch_adapt}
\end{equation}

Corbacho et al. (2005b) followed a particular neural network
implementation of both mappings $M^{x,y}_P (t)$ and $T_P (t)$ by a learning spatio-temporal mapping algorithm.
Sanchez-Montanes and Corbacho (2004) presented an information theoretic metric to build this type of mappings.

Associated to the predictive schema $P^{x,y}$ there is always its corresponding dual schema $S^{y,x}$.
The predictive schema is the analogous to the forward internal models and the dual schema is the analogous to the inverse internal model (Jordan \& Rumelhart, 1992).
\\
\\
{\bf Definition:} A {\it dual schema} $S^{y,x}$ associated with predictive schema $P^{y,x}$, 
produces the necessary optimal pattern of activity input to the cause schema $S^y$ 
in order to successively achieve the desired goal pattern of activity in the effect schema $S^x$.
It is a schema with the following special characteristics:

$<Iplist>$: The first input port is of the same type and connects to the
output port of the effect schema $S^x$, i.e. $i^{y,x}_1(t)=o^{x}(t)$.
The second input port is of the same type and connects to the
output port of the goal schema $G^{z,x}$, i.e. $i^{x,y}_2(t)=o^{*z,x}(t+1)$,
that is, a goal schema that is in charge of producing a goal pattern of activity for the effect schema $S^x$ 
(defined below).
The remaining input ports are optional and correspond to {\it context}
 information; i. e. $i^{y,x}_3(t)=o^{v}(t)$

$<Oplist>$: The output port is of the same type and connects to the input port of cause schema $S^y$ 

$<Varlist>$: Includes the parameters of the dual schema, namely ${\bf W}^{y,x}$.	

$<Behavior>$: The dual schema behavioral specification includes a mapping $M^{y,x}$ parameterized by ${\bf W}^{y,x}(t)$, such that
\begin{equation} 	
\o^{y,x}(t)=M^{y,x}(o^{x}(t), o^{*z,x}(t+1), o^v(t); {\bf W}^{y,x}(t) )
\label{eq_dual_sch}
\end{equation} 
as well as a mapping $T$, 
\begin{equation}
{\bf W}^{y,x}(t+1)=T({\bf W}^{y,x}(t), 
 o^{x}(t+1), 
 o^{*z,x}(t+1),
\frac{\partial \hat{o}^{x,y}}{\partial o^{y,x}},
\frac{\partial {o}^{y,x}}{\partial {\bf W}^{y,x}})
\label{eq_dual_adapt_sch}
\end{equation}
$T$ allows the dual schema to be tuned, i.e. its parameters 
change according to the performance error $(o^{*z,x}(t+1) - o^{x}(t+1))$,
so that the observed response $o^{x}(t + 1)$
becomes increasingly closer to the desired goal response $o^{*z,x}(t + 1)$ as the number of interactions increases. 
The problem of training the dual schema is a  {\it distal learning} problem (Jordan \& Rumelhart, 1992) since the parameters of the dual schema must be adapted based on the error on a distal space. That is, the dual schema must find parameters that recover the optimal patterns in the space of the cause schema, that is in the representation space of  $o^{y}(t)$ 
so as to reduce the difference in the distal error space $(o^{*z,x}(t+1) -  o^{x}(t+1))$. 
As Jordan and Rumelhart (1992)  described, and online learning algorithm based on stochastic gradient descend can be used. 

To perform adaptation, the change of weights must take into account  $\frac{\partial o^{x,y}}{\partial o^{y,x}}$. 
Yet, the dependence of $o^{x,y}$ on $o^{y,x}$ is assumed to be unknown a priori. 
Yet, given a differentiable predictive forward model, it can be approximated by $\frac{\partial \hat{o}^{x,y}}{\partial o^{y,x}}$. That is, the distal error is propagated backward though the predictive schema (forward model) and down into the dual schema (inverse model) where the weights are actually changed accordingly.
For the sake of clarity and simplification for this article, we assume that only one predictive and its corresponding dual schema are instantiated for each specific active motor schema. Hence,  avoiding issues of integration (linear on nonlinear). Thus, $\hat{o}^x(t) = \hat{o}^{x,y}(t)$, $o^y(t) = o^{y,x}(t)$ and $\frac{\partial \hat{o}^x}{\partial o^y} = \frac{\partial \hat{o}^{x,y}}{\partial o^{y,x}} $.
In this regard, equation 19 will also be simplified to reflect this assumption.
Kawato's (1990a) feedback-error-learning can also be used in this context. 
Later, we will schematize this information flow once a shema-based model is introduced and described.

\subsection{Goal schemas and goal-oriented behavior}
Goal oriented behavior is one of the hallmarks of intelligent systems.
That is, the ability to set and achieve a wide range of goals
 (desirable states defining an objective signal).
Goal states must be stored so that they can be actively sought for in the future.
We must distinguish between implicit (hardwired) goals and learned goals and subgoals.
Drive reduction pertains to the first kind.
Where {\bf drives} can be viewed as states (Milner, 1977), which influence neurons either
 mechanically or chemically, and which have representations. Based on this hypothesis,
Arbib \& Lieblich (1977)
introduced a set $(d_i,...)$ of discrete drives to control the agent behavior. 
At each time $t$, each drive $d_i$ has a value $d_i(t)$. Drives can be appetitive or aversive.
Each appetitive drive spontaneously increases with time towards $d_{max}$, while
 aversive drives are reduced towards 0, both according to a factor $\alpha_d$. 
An additional increase occurs if an incentive $I(d,x,t)$ is present such as the sight or aroma of food
 (e.g. $S^{PREY}$ schema active) in the case of hunger.
Drive reduction $a(d,x,t)$ takes place in the presence of some
substrate -food reduces the hunger drive. If the agent is in the situation $x$ at time $t$, then the value of $d$ at time $t+1$ will become

\begin{equation} 
d(t+1)=d(t)+\alpha_d|d_{max}-d(t)|-a(d,x,t)|d(t)| + I(d,x,t)|d_{max}-d(t)|
\end{equation} 
\\
Internal drives (variables) must be kept within a restricted interval to assure
the survivability of the agent (intrinsic hardwired goals).
A particular state $x(t)$ becomes a goal state $x(t)^*$ if it is directlly associated to drive reduction $a(d,x,t)$ or is predictive of drive reduction.
From this a hierarchy of subgoals has to be learned. That is, what states take the 
system to the primary goals (Guazzelli, Corbacho, Bota \& Arbib, 1998). 
Corbacho (1997) included $hunger$, $fear$, $thirst$, etc; 
in this paper we have included just $hunger$.
A goal corresponds to a state with a high drive reduction or in the way 
(anticipating drive reduction).
$x$ in the definition above is the state of the output port of a schema (or schemas)
at time $t$.
Corbacho (1997) described reinforcement type algorithms to store primary sensory goal states, that is, sensory states associated with high reward or anticipation of reward. 
These are specially needed in stochastic environments with delayed reinforcement (Sutton, 1988; 1990; Sutton \& Barto, 1998), more details in Corbacho (1997).

Hence, the ability to set and generate goals and subgoals is critical, such as the goal state of the jaw muscle  spindles indicating that the mouth must be open in order to get the prey. 
So the constructive brain architecture must be able to restore desired states so that they can be actively pursued. 
The desired state in a particular schema is triggered by the contextual state defined by the activity pattern in another schema. 
For secondary goals, goal states must be parametrized by contextual information.
That is, the goal pattern of activity must be produced by an adaptive mapping. 
So, for instance, prey-catching reduces the hunger drive,
which is signaled by the prey in the mouth,
so that when a prey is within the visual field a subgoal must be generated so that the jaw muscle spindles must get activated, indicating that the mouth is successefully open in order to allow the prey to get into the mouth.
During leaning, the reinforcement signals enhance the elegibility of the projections from the prey recognition schema to the jaw spindles goal pattern of activity representation (Corbacho, 1997). 
Another example in adaptive navigation corresponds to the goal schema  that 
produces a goal pattern of representation in the motor heading map in the presence of a target, namely representing that in order to capture the target it must be centered within its sensorimotor representation, that is the target is "within grasp". 
Sections  6.2.1 and 7.3.1 will detail these goal schemas learned during learning to detour and learning to grip in respectively. 
\\
\\
{\bf Definition:} A {\it goal schema} $G^{z,x}$ associated to (source) schema $S^z$ 
and (objective) schema $S^x$ (typically an effect schema in a dual/predictive schema) 
is a schema with the
following special characteristics:

$<Iplist>$: The first input port is of the same type and connects to the 
output port of the source schema $S^z$, i.e. $i^{z,x}_1(t)=o^{z}(t)$.
The remaining input ports are optional and correspond to {\it contextual}
 information.

$<Oplist>$: The output port contains the {\it objective} response
${o}^{* z,x}(t+1)$, representing the desired pattern of activation in $o^{x}(t+1)$.  	

$<Varlist>$: Includes parameters of the goal schema, ${\bf {W}}^{z,x}_G$.	
It also includes the matrix $V$ for value maximization by reinforcement learning type dynamics that selects goal patterns of activity 
as well as internal variables $G$ to store temporal delayed states:

$<Behavior>$: Implements a mapping $M^{z,x}_G$ from different inputs at different
times to the objective response, 
\begin{equation} 	
{o}^{* z,x}(t+1)=M^{z,x}_G( o^{z}(t), G(t)   ; {\bf W}^{z,x}_G (t) )
\end{equation} 
where
\begin{equation}
G(t - 1) := o^{*z,x} (t), ... ,\\
G(t - n) := G(t -n- 1)
\end{equation}
and := corresponds to the assignment operator. \\
\noindent It also includes  a mapping $T_G$,
\begin{equation}
\begin{split}
{\bf W}^{z,x}_G (t+1)=T_G({\bf {W}}^{z,x}_G (t),
{o}^{* z,x}(t+1), 
o^{x}(t+1))
\end{split}
\end{equation}
that attempts to reduce the difference between ${o}^{* z,x}(t+1)$ and $o^{x}(t+1))$, 
adapted by the training set selected by value maximization selecting only  paterns  $( o^{*z,x}_\alpha(t_i), o^x_\alpha(t_i))$ that maximize expected value, that is, 
selected in case at time $t_i$: 
\begin{equation} 
V(o^{*z,x}_\alpha(t_i)), t_i) > \beta
\end{equation} 
where
\begin{equation} 
V(x, t) =e(t) + \epsilon \sum_{i,j} c_{x,y} \cdot V(y, t+ 1)
\end{equation} 
$e(t)$ corresponds to direct reinforcement associated to drive reduction, that is $e(t) = a(d,x,t)$,
for instance $target\_in\_grip(t)$ receives inmediate reward associated to the hunger drive reduction, $c_{x,y}$ corresponds to the matrix of goal states dependencies learned by cause-effect type dynamics (section 4.1) and 
$\epsilon < 1$. 
For instance, only certain subset of  $grip(t)$ patterns achieve a high enough value.
The value is received since it is time correlated with other states of high value. 
These high value grip patterns are the only ones that are learned in relation to the target pattern. 
This is a type of reinforcement learning (Sutton, 1988; 1990; Sutton \& Barto, 1981; 1998) and has been shown to explain anticipatory neural activity (Suri \& Schultz, 2001). 
\\
\\
A specific spatio-temporal mapping learning algorithm to learn this mapping is
presented in (Corbacho et al., 2005b) and a generalized information-theoretic measure is presented in (Sanchez-Montanes \& Corbacho, 2004).

\section{Self-constructive pseudo-Algorithm and Dynamics}

As previously described in section 1.3.1 the initial agent functionality/structure (at time $t_0$)  is predefined by the seed schema(s) structure determined by the tuple:
\begin{equation}
A(t_0) = (D,  S(t_0)= \{S^1, ..., S^N\}, C(t_0), Q(t_0), R(t_0), L)
\end{equation}
$S(t_0) = \{S^1, ..., S^N\}$ determines the set of initial (pre-wired) seed shemas $S^1, ... , S^N$;
$C(t_0)$ determines the initial port-to-port connection map; $Q(t_0)$ determines the initial schema support network; $R(t_0)$ determines the cause-effect relations between different schemas, typically initially empty since they are learned by observing the effects of the interactions after they have been experimented by the agent.
$L$ will incrementally extend and adapt $S(t)$, $C(t)$, $Q(t)$, and $R(t)$ accordingly in order to 
maximize $A$ coherence with its environment. 
This section provides a schematic description of the pseudo-algorithm by describing the different algorithmic components, namely:
\\
\\
\indent A) Schema dynamics:
\\
\indent B) Value, Performance and Coherence maximization for Schema Adaptation
\\
\indent C) Schema Construction conditions
\\
\indent D) Schema Construction
\\
\\
This different algorithmic components are activated following a cooperative and competitive distibuted model of computation (Arbib,1976).
Hence, giving rise to a dynamic configuration of these components 
This section provides a schematic description of the pseudo-algorithm. Later sections describe in more detail some 
important aspects of the algorithm. Section 4.1 describes in more detail the Cause-Effect dynamics and section 4.2
provides a more detailed description of the Schema Construction process central to the SCAI algorithm.

\newpage
\begin{bf}A) Schema dynamics \end{bf}
\\

\begin{center}
\begin{doublespace}

\begin{tabular}{| l |  }
\hline
:::::::: \begin{bf} a) Drive dynamics and primitive goal generation   \end{bf}:::::::	\\
\hline
$\forall  d_i \in D$:	
$  d_i(t+1)=d_i(t)+\alpha_i|d_i^{max}-d_i(t)|-a(i,t)|d_i(t)| 
+I(i,t)|d_i^{max}-d_i(t)|    $ \\
\hline
\end{tabular}
\end{doublespace}
\end{center}
\begin{center}
\begin{doublespace}

\begin{tabular}{| l |  }
\hline
:::::::: \begin{bf} b) Primitive perceptual schema assertion    \end{bf}:::::::	\\
\hline
$\forall i, j$ s.t. $E(o_j)(t)  \neq \sharp$: $input^i(t) = M^i(E(o_j))$ \& $a^i(t) = \Theta^i + \delta$ 
 \\
\hline
\end{tabular}
\end{doublespace}
\end{center}
\begin{center}
\begin{doublespace}

\begin{center}
\begin{doublespace}

\begin{tabular}{| l |  }
\hline
:::::::: \begin{bf} c) Schema support propagation:   \end{bf}:::::::	\\
\hline
$\forall  i $:	
$a^i(t+1) = \sigma (\sum_j q_{i,j}(t)  \cdot a^j(t))$\\
(for details on the different constraints on $q_{i,j}$ refer to Corbacho \& Arbib (2019)\\
\hline
\end{tabular}
\end{doublespace}
\end{center}

\begin{center}
\begin{doublespace}

\begin{tabular}{| l |  }
\hline
:::::::: \begin{bf} d) Subset goal schema assertion: \end{bf}:::::::	\\
\hline
$\forall a^z(t)$ such that $a^z(t) > \Theta^z$,  $o^z(t) \neq \sharp$ \\
$\o^{* z,x}(t+1)=M^{z,x}_G( o^{z}(t), G(t)   ; {\bf W}^{z,x}_G (t) )$\\
where
$G(t - 1) := o^{*z,x} (t), ... ,
G(t - n) := G(t -n- 1)$\\
\hline
\end{tabular}
\end{doublespace}
\end{center}

\begin{center}
\begin{doublespace}

\begin{tabular}{| l |  }
\hline
:::::::: \begin{bf} e) Subset Dual and Predictive schema assertion \end{bf}:::::::	\\
\hline
$\forall a^{*z,x}(t), a^v(t)$,  such that $a^{*z,x}(t) > \Theta, a^v(t) > \Theta$,  $o^{*z,x}(t) \neq \sharp$	\\	
$o^{y,x}(t)=M^{y,x}(o^{x}(t), o^{*z,x}(t+1), o^v(t); {\bf W}^{y,x}(t))$ \\
\hline
and if $a^{y,x}(t) > \Theta$,   $o^{y,x}(t) \neq \sharp$		\\
$\hat{o}^{x,y}(t+1) = M^{x,y}_P (o^{x}(t), o^{y,x}(t), o^{v}(t); {\bf W}^{x,y}_{P} (t))$ \\
\hline
\end{tabular}
\end{doublespace}
\end{center}

\begin{tabular}{| l |  }
\hline
:::::::: \begin{bf} f) Primitive motor schema assertion    \end{bf}:::::::	\\
\hline
$\forall  i, j$ s.t. $output^i(t)  \neq \sharp$ \& $a^i(t) > \Theta^i$:   $E(i_j)(t) = output^i(t)$
 \\
\hline
\end{tabular}
\end{doublespace}
\end{center}

\newpage

\begin{bf} B) Value, Performance and Coherence Maximization for Schema Adaptation \end{bf}

\begin{center}
\begin{doublespace}

\begin{tabular}{| l |  }
\hline
:::::::: \begin{bf}  a) Adapt goal schemas: value maximization and training goal mappings\end{bf}:::::::	\\
\hline
$V(x, t) =e(t) + \epsilon \sum_{i,j} c_{x,y} \cdot V(y, t+ 1)$\\
\hline
$\forall z,x$ if 
$V(o^{*z,x}_\alpha(t_i), t_i) > \beta$ then \\
${\bf W}^{z,x}_G (t_i+1)=T_G({\bf {W}}^{z,x}_G (t_i),
{o}^{* z,x}_\alpha(t_i), 
o^{x}_\alpha(t_i))$\\
\hline
\end{tabular}
\end{doublespace}
\end{center}
\begin{center}
\begin{doublespace}

\begin{tabular}{| l |  }
\hline
:::::::: \begin{bf} a') Adapt  dual \& predictive schemas parameters by distal learning:  \end{bf}:::::::	\\
\hline
if 	 $(o^{*z,x}(t+1) - o^{x}(t+1)) > \epsilon$ 	\\	

${\bf W}^{y,x}(t+1)=T({\bf W}^{y,x}(t), 
 o^{x}(t+1), 
 o^{*z,x}(t+1),
\frac{\partial \hat{o}^{x,y}}{\partial o^{y,x}},
\frac{\partial {o}^{y,x}}{\partial {\bf W}^{y,x}})$ \\
\hline
if	 $(o^{x}(t+1) - \hat{o}^{x,y}(t+1)) > \epsilon$ 		\\
${\bf W}^{x,y}_P (t+1)=T_P ({\bf W}^{x,y}_P (t), 
\hat{o}^{x,y}(t+1), 
o^{x}(t+1), 
\frac{\partial \hat{o}^{x,y}}{\partial {\bf W}^{x,y}_P }, )$ \\
\hline
\end{tabular}
\end{doublespace}
\end{center}

\begin{bf} C) Performance and Coherence Maximization conditions for Schema Construction \end{bf}

\begin{center}
\begin{doublespace}
\begin{tabular}{| l |  }
\hline
:::::::: \begin{bf} Unexpected event, Goal and Context relations \end{bf}:::::::	\\
\hline
Unexpected incoherence conditions:    \\

a) $\exists x, y$  such that $o^x(t+1) = \sharp$   \&  $\hat{o}^{x,y}(t+1) \neq \sharp$, or \\

a') $\exists x, y$  such that $o^x(t+1) \neq \sharp$   \&  $\hat{o}^{x,y}(t+1) = \sharp$ 	
 \\
\hline
b) Approaching goal:    
\\

 $\exists x, y$  such that $d(o^{*z,x}(t + 1), o^x(t + 1) )    \le     d(o^{*z,x}(t+1),  o^x(t))$ 
\\
b') Away from goal:    
\\
$\exists x, y$  such that $d(o^{*z,x}(t + 1), o^x(t + 1) )    >     d(o^{*z,x}(t+1),  o^x(t))$ 
\\
\hline

c) Unexpected new cause-effect relation: 
\\
$\exists x, y$  such that $(S^x, S^y)$ where $R_{x,y} > \Theta$	and $R$ dynamics are described in section 4.1 
\\
\hline
d) Define the context by the schema(s) $S^v$:\\
active before, during and after 
the unexpected event  takes place at $t$: \\
$o^v[t-n\delta, t +n\delta] \neq \sharp$
\\
\hline
\end{tabular}
\end{doublespace}
\end{center}

\newpage
\begin{bf} D) New schema construction and cause schema re-construction\end{bf} 

Unexpected event (C.a or C.c) and approaching the goal (C.b) have selected the schemas triplet indexed by $[x,y, v]$  
which serves to fill in the slots for the new to be constructed schemas. 
\\

a) Recruitment of overgeneralized schema dormant structures: 

Predictive schema and Dual schema. Recruit from array of indexed functions:
\begin{equation}
S^{N+1} (i^1, \dots i^k ; o^1, \dots,  o^m);  \\ 
P^{N+1} (i^1, \dots i^k ; o^1, \dots,  o^m)
\end{equation}

\begin{center}
\begin{doublespace}

\begin{tabular}{| l |  }
\hline
:: \begin{bf}  
Instantiation of the recruited overgeneralized schema structures    \end{bf}::	\\
\hline
b) New predictive schema: $P^{N+1} = P^{x,y,v}$
\\
\indent \indent \indent $i^{x,y}_1(t) = o^x(t)$
\\
\indent \indent \indent $i^{x,y}_2(t) = o^y(t)$
\\
\indent \indent \indent $i^{x,y}_3(t) = o^v(t)$ 
\\
\indent \indent \indent $\hat{o}^{x,y}(t) = E[o^y(t+1) | o^x(t), o^y(t), o^v(t)]$
\\
\hline
	b') New dual schema: $S^{N+1} = S^{y, x, v}$
\\
\indent \indent \indent $i^{y,x}_{1}(t) = o^x(t)$ 
\\
\indent \indent \indent $i^{y,x}_{2}(t) = {o}^{*z,x}(t + 1)$
\\
\indent \indent \indent  $i^{y,x}_3(t)=o^{v}(t)$
 \\
\hline
b'') New input port to re-construct cause schema $S^{y'}$:\\
$i^{y'}_{k+1}(t) = o^{y,x}(t)$; 
\\
\hline
\end{tabular}
\end{doublespace}
\end{center}
\[
  o^{y'}(t) = \begin{cases}
    i^{y'}_{k+1}(t)  & \text{if $i^{y'}_{k+1}(t) \neq \sharp$} \\
   o^y(t) & \text{otherwise}
  \end{cases}
\]

The schema set changes to $St+1) = S(t) \cup \{S^{y, x, v}, P^{x, y, v}, S^{y'}\} - S^y$
\\
\\
In the next subsection the schema construction process will be described in more detail given the relevance of this operator in the SCAI framework.
Next, we will introduce the mechanisms to learn the effective 
cause-effect relations within a complex system (e.g. brain, autonomous agent).

\newpage

\subsection{Constructing Cause-Effect relations}
In this section we propose learning mechanisms that will be able to
construct the cause-effect relations from the combinatorially
large relational space.  
As already indicated, these cause-effect relations will in turn allow for mental simulation and self-experimentation.
To decide on a cause-effect relation between
two schemas $S^x$ and $S^y$ we will use some measure
of {\it delayed similarity} of the activity patterns out of their
output ports during a time interval i.e. $o^x[t_a,t_b]$ and 
$o^y[t_a+\tau,t_b+\tau]$ respectively.
In general we will assume that any schema connects to any other
schema through several
connections, we assume that these different connections will have different
"delay properties", thus giving rise to different responses in
different connections. 
We now define the instantaneous degree of cause-effect relation between two patterns
$o^x(t)$ and $o^y(t)$, at a point in time $t$ by the
following measure

\begin{equation}
c^{x,y}_{\tau} (t) = \Theta[o^x(t)] \cdot \Theta[o^y(t)]-\alpha d^{x,y}_{\tau}(t) 
\end{equation}
\noindent where  
\begin{equation}
d^{x,y}_{\tau}(t) = (o^y(t-\tau) - o^x(t))^2
\end{equation}
\[
  \Theta[x] = \begin{cases}
    1 & \text{if $x \neq 0$} \\
    0 & \text{otherwise}
  \end{cases}
\]

Two signals may not be very similar in their fine temporal structure
yet they might be similar in their qualitative behavior, that is,
when one is active the other is also active, when one is large the
other is large, when one is growing the other is growing, and so
on. If one of the signals is inactive and the other is active the
connection is decreased, so we implement a sort of AND operation, that is,  if both are active
the first term in equation 12 becomes 1, hence increasing the weight in
     the first case and decreasing the weight otherwise. At
     the end only those that maintain a recurrent and constant delay
     relation remain. The threshold factors avoid correlations among layers
     that are both inactive. Since when both are inactive the first
     term becomes 0 as both its factors are 0 and the second term is
     also 0. On the other hand, if one is active and the other is
     inactive the "distance" of both signals becomes $\alpha
d^{x,y}_{\tau}(t)$.
Finally if both are active their "similarity" becomes $1-\alpha d^{x,y}_{\tau}(t)$.

The second phase consists of calculating how {\it reliable} the
instantaneous cause-effect relation is within a time interval $[t_0, t_n]$. 
This is calculated by

\begin{equation}
r^{x,y}_{\tau} (t+1) = r^{x,y}_{\tau} (t) + \beta c^{x,y}_{\tau} (t)
\end{equation}
\noindent so that at the end of the training period the measure of the
similarity of both signals will be approximated
by $r^{x,y}_{\tau} (t_N) = \beta\sum_{t=t_0}^{t_n}c^{x,y}_{\tau}(t)$
When a correlation is not "reliable" it will be positive in certain
occasions and negative in others, thus, overall the reliability
measure will be close to 0. From now on ${\bf r}$ denotes the matrix with
components $r^{x,y}_\tau$.
The delays between actions and sensory feedback are reflected on ${\bf
r}$. 
When a cause-effect correlation is reliable (above a
certain threshold) the cause schema becomes the schema that outputs  $o^y(t)$, namely $S^y$.
and the "effect"schema becomes the schema that outputs $o^x(t)$,
namely $S^x$.
Hence, giving rise to the construction of a new predictive schema $P^{X,Y}$.
SBL also constructs its corresponding dual schema $S^{Y,X}$ in parallel to be able to re-construct a previous successful pattern of interaction.

\begin{figure}[ht!]
\centering
\includegraphics[width=100mm]{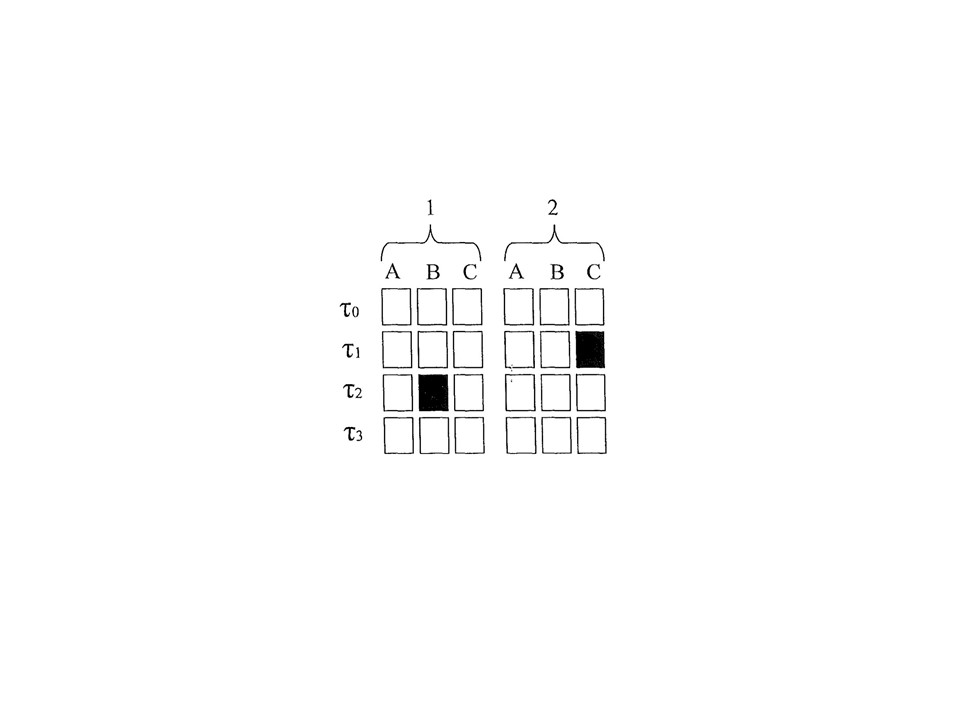}
  \caption{{\bf  Representation of the cause-effect relations} in the simplified example consisting of five schemas, 
three of which are motor schemas A, B, C and two of which are perceptual schemas 1, 2. 
The black squares represent the existence of a cause-effect relation between the respective motor and perceptual schemas with that specific delay $\tau_i$.
On the other hand, the white squares represent the lack of any cause-effect relation between the respective motor and perceptual schemas. }
  \label{CE}
\end{figure}

To facilitate understanding let us first analyze a simplified case.
Consider five schemas, three of which are motor schemas A, B, C and two of which are perceptual schemas 1, 2. 
Suppose there is only a cause-effect relation between motor schema B and perceptual schema 1 with a delay $\tau_2$; 
and a cause-effect relation between motor schema C and perceptual schema 2 with a delay $\tau_1$.  
Figure \ref{CE} represents the cause-effect relations among these three motor schemas and these two perceptual schemas. 
To farther exemplify suppose that perceptual schema 1 is sensory feedback from the grip and 
perceptual schema 2 corresponds to "target on view field". 
On the other hand, suppose that motor schema A is sidestep,  motor schema B corresponds to ´open grip´, and motor schema C corresponds to "lunge". 
It should be clear in this example  that there is only a cause-effect relation between the schemas "open grip" 
and sensory feedback from the grip with a certain time delay; and the schemas sidestep and "target on view field" with a probably different time delay.

\subsection{Schema Construction: open-ended self-growing}
\begin{figure}[ht!]
\centering
\includegraphics[width=120mm]{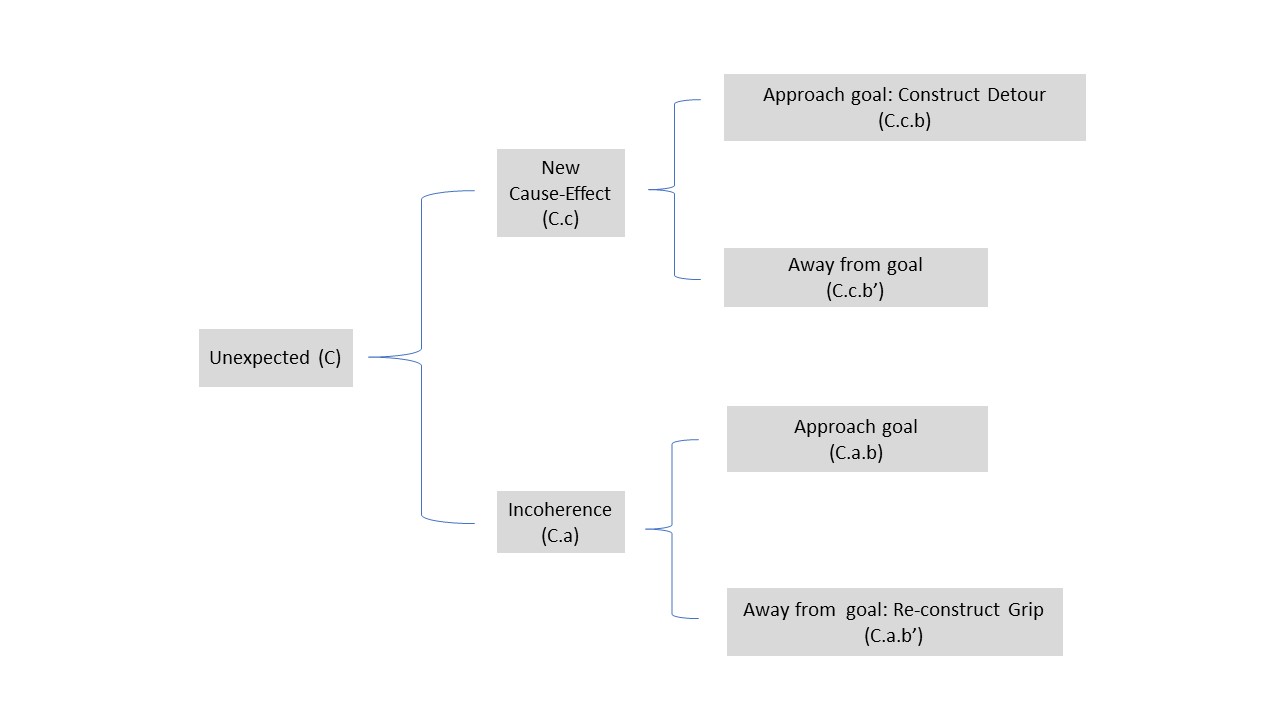}
  \caption{{\bf  Unexpected can be generated in two cases: }(C.c) by a new predictive cause-effect relation discovered by the cause-effect dynamics (as described in section 4.1); 
and (C.a) when there is an incoherence, that is, the predictive response does not match the actual observed response
$(o^x(t+1) - \hat{o}^{x,y}(t+1)) \neq 0$
and, at the same time, either the cause or the effect schemas are inactive,
i.e., $o^y(t) = \sharp$ or $o^x(t+1) = \sharp$.  }
  \label{ConstructionConditions}
\end{figure}

In general, to construct a new schema, three aspects must be determined, namely: the trigger event (when), the components (what), and the topological configuration (how).
First of all, the agent must realize the trigger event when there is the need to construct (a) new schema(s). 
In this regard, both the prediction error (incoherence) and the performance error will serve as the triggers under specific conditions, as it will be explained below. 
The next step consists of realizing what can take the system back to coherence and closer to the goal.
This implies determining what components and in which topological configuration will make up the new schema.
As already explained, SBL attempts to maximize coherence, that is, reduce the prediction error, and, at the same time, maximize performance. 
In this regard, structural learning processes are triggered when specific unexpected patterns of interaction occur, either due to a new cause-effect relation or to a an incoherence. Figure \ref{ConstructionConditions} details the different conditions that may take place. 

An unexpected event can be associated with the agent getting closer (C.b) or getting away from a specific goal (C.b').  
In the first case (C.b), the system should attempt to "record" the "configuration" that gave rise to that interaction, such as is the case in learning to detour that will be explained in section 6.5 in Paper II of this series. 
Whereas in the case (C.b'), the system must make sure to avoid
the current pattern of interaction and re-construct some previous successful pattern (or construct alternative strategies (Corbacho, 1997)). Learning after failure for break-out corresponds to this case 
since the failure causes a prediction error due to the inactivation of the effect schema. 
In this case, previous successful pattern re-construction is necessary to achieve the goal as it will be explained in section 7.4 in Paper III in this series. 
Other cases in the figure are included in Corbacho and Arbib (2019). 
\\
\\
{\bf Constructing new predictive schema $P^{x,y, v}$:} 
As already explained in section 3.2, once the effect $S^x$ and the cause schema $S^y$ have been determined, the predictive schema behavioral specification is defined by a mapping $M^{x,y}_P$ parametrized by ${\bf W}^{x,y}_{P} (t)$, as defined in equation 1.
The new predictive schema recruits the dormant structure and parametrizes 
${\bf W}^{x,y}_{P} (t)$ accordingly.
Next we define the general constructive equation.
\\
\\
\indent \indent \indent $i^{x,y}_1(t) = o^x(t)$
\\
\indent \indent \indent $i^{x,y}_2(t) = o^{y,x}(t)$
\\
\indent \indent \indent $i^{x,y}_3(t) = o^v(t)$
\\
\indent \indent \indent $\hat{o}^{x,y}(t+1) = E[o^{x}(t+1) | o^{x}(t), o^{y,x}(t), o^{v}(t)]$
\\
\\
where adaptation of ${\bf W}^{x,y,v}_P$ is driven by the minimization of $d(o^x(t+1),  \hat{o}^{x, y}(t+1))$. 
Also its dual schema $S^{y,x}$ gets constructed in parallel.
The dual schema requires as input a goal pattern that must be  provided by a goal schema.
In this regard, the goal schema $G^{z,x}$ learned in
previous successful interactions outputs the "goal" activity
pattern for schema $S^x$ as already explained in section 3.3. 
\\
\\
{\bf Constructing new dual schema $S^{y,x,v}$:} 
This new schema's role is to
produce the "optimal" pattern of activity in the cause schema $S^y$ that will in turn give rise, 
though the cause-effect dynamics, to the goal pattern of activity in the effect schema $S^x$  
The components of the new schema $S^{y,x}$ are as follows:
\\
\\
\indent \indent \indent $i^{y,x}_{1}(t) = o^x(t)$ 
\\
\indent \indent \indent $i^{y,x}_{2}(t) = {o}^{*z,x}(t + 1)$
\\
\indent \indent \indent  $i^{y,x}_3(t)=o^{v}(t)$
\\
\\
\indent  $<Oplist>$: the output port is of the same type as $S^{y}$ input port
and connected to the re-constructed cause schema $S^{y'}$ (defined below).
\\
\indent $<Behavior>$: the system must produce a modulatory activity pattern input to the re-constructed cause schema  $S^{y'}$ (defined below) such that it will make the pattern $o^x(t)$ in the effect schema $S^{x}$ get closer to the goal pattern ${o}^{*z,x}(t+1)$ set by the goal schema $G^{z,x}$.
Where adaptation of ${\bf W}^{y,x, v}_P$ is driven by the minimization of $d(o^{*z,x}(t +1 ), o^{x}(t + 1))$. 
This is analogous to learning the inverse model in distal supervised learning (Jordan \& Rumelhart, 1992). 
\\
\\
The cause schema $S^y$ receives a new modulatory input from the dual schema $S^{y,x}$
since the dual schema outputs an optimal pattern of activity for $S^y$  under a specific context.
A particular context conditions a specific optimal pattern (Torralba, 2003). 
These modulatory loops have been shown to have already evolved in lower vertebrates (Ewert, et al., 2006). 
Hence, the cause schema is reconstructed to include this new modulatory input. 
\\
\\
{\bf Reconstruct Cause Schema $S^y$ into $S^{y'}$:}  
Cause schema $S^y$ is reconstructed to incorporate
a new modulatory input port from the newly constructed dual schema (represented by dashed lines in Figures 9 and 10). 
In turn, it reconstructs its
behavioral specification as now it has to take into account this new instantiated
modulatory signal. 
\\
\indent $<Iplist>$: They are the same as in $S^y$ (i.e. k input ports) in
addition to a newly instantiated modulatory input port of the same
type and connected to the output port of  the dual schema $S^{y,x}$,  namely,
\begin{equation}
i^{y'}_{k+1}(t) = o^{y,x}(t)
\end{equation}
\indent $<Behavior>$: The behavioral specification remains the same 
as in  $S^y$  but two extra
assignment expressions are instantiated at the very end of the behavioral
specification to take into account the new modulatory input signal
that serves to reconstruct the pattern of activity proven to be successful in the
past in similar conditions, namely,   
\[
  o^{y'}(t) = \begin{cases}
    i^{y'}_{k+1}(t)  & \text{if $i^{y'}_{k+1}(t) \neq \sharp$} \\
   o^y(t)  & \text{otherwise}
  \end{cases}
\]
That is, if there is no modulatory input, the output port assignment remains as in the original schema $S^{y}$. Yet, if there is any modulatory input, then the output port assignment corresponds to the input objective modulatory input. As already expressed, for the sake of clarity and simplification for this article, we assume that only one predictive and its corresponding dual schema are instantiated for each specific active motor schema. Hence,  avoiding issues of integration (linear or nonlinear). Thus, $\hat{o}^x(t) = \hat{o}^{x,y}(t)$, $o^y(t) = o^{y,x}(t)$ and equation 12 can be simplified to reflect this assumption.

\begin{figure}[ht!]
\centering
\includegraphics[width=120mm]{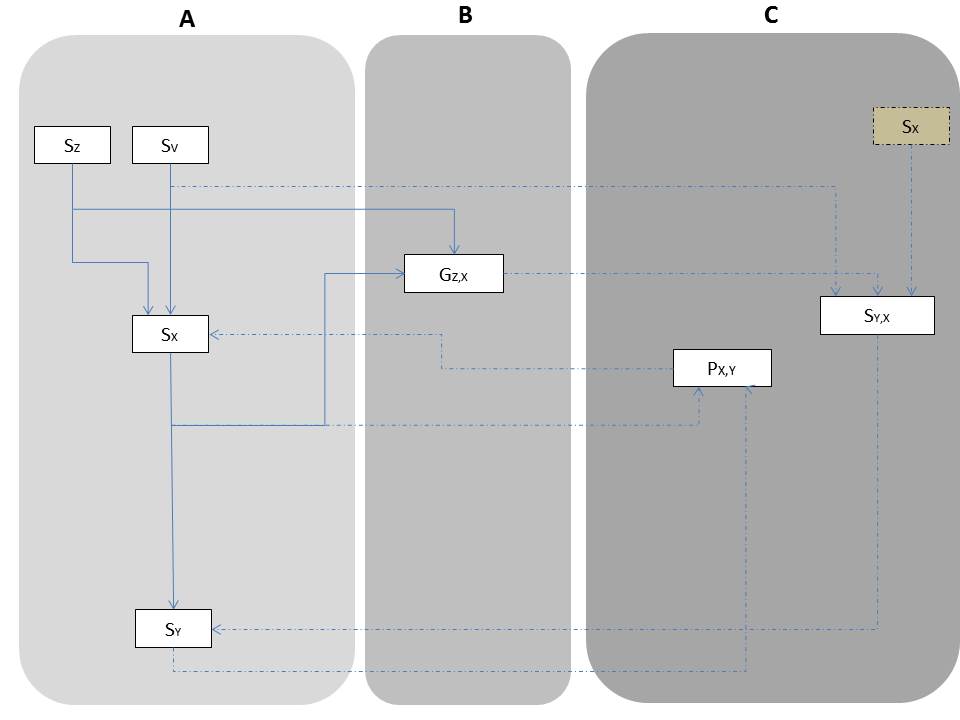}
  \caption{{\bf Generalized process of schema(s) construction}. 
Construction of the new  predictive schema  $P^{x,y}$ and its dual schema $S^{y,x}$.
In turn, cause schema $S^y$ is adapted to instantiate
a modulatory input port from the dual schema (represented by dashed lines).
This in turn adapts its
behavioral specification as now it has to take into account this instantiated
modulatory signal. }
  \label{fig_construct}
\end{figure}

In summary, a global picture of this process of schema construction  is depicted in
Figure \ref{fig_construct}. It includes the predictive schema  $P^{x,y}$ and its associated dual schema $S^{y,x}$,
as well as the involved goal schema $G^{z,x}$.

\section{Conclusions}

This paper presents self-constructive AI as a general framework that allows for the autonomous construction of increasingly functionality in AI systems.
Hence, $SCAI$ is one possible architecture for Artificial General Intelligence since it is capable to develop qualitative new different behaviors not previously imagined by its designers. In order to do so, it incorporates structural learning which allows the construction of new structures (schemas) based on previous structural components (schemas). 
The general idea behind the $SCAI$ framework is to start with less pre-designed structures and functions and more powerful and general constructive mechanisms so that autonomous development can take place. 

Self-constructive can be achieved by three central principles of organization, namely:
Self-growing implies that the system autonomously and increasingly is able to construct new useful patterns of interaction with its environment and reflecting those in new internal representations. 
Self-experimental: the agent can also perform experiments with the environment and also internal experiments (mental simulation) with many previously constructed models that allow to predict and anticipate the results of its actions, thus allowing to improve its decision making by matching the expected results with the desired results. 
Finally self-repairing implies that the agent is able to recover from possible break-outs in any of its functionality by constructing internal models that allow to model and later re-construct previously successful patterns of interaction. Hence, giving rise to more flexible and robust agents.

We propose $SBL$ as one possible formalization for the $SCAI$ constructive architecture. 
$SBL$ allows for the construction of predictive (cf. forward internal models) and its dual schemas as active processes capturing relevant patterns of interaction; and we suggest that intelligence  emerges from myriads of these patterns as well as their dual predictive internal models associated.
This paper also provides a pseudo-algorithm for $SCAI$.
For a framework to be general it must include perception, sensory integration, decision making, action and adaptation, 
which must be intrinsic to all the previous components. 
Integration from multisensorial to sensorimotor transformations is critical. 
Decision of (near) optimal actions must not only take into account the current inflow of information but  
it must also simulate the expected outcome of the different possible actions in order to select (and plan) the most valuable one. 
Also, for perception to be general, it must be able to cope with different kinds of receptors, explore and discover different  manifolds in receptor space, different relations, measure their value, store them and relate them to expectations. 
Similarly, for action  to be general, it must be able to cope with different kinds of effectors, explore and discover different  manifolds in effector space, different spatiotemporal relations, measure their value, store them and relate them to expectations. 
Finally, for an AI framework to be general, it must allow for different tasks within the same environment (e.g. incremental navigation) as well as completely different environments and domains of interaction. 
To demonstrate the generality of the $SCAI$ framework we provide three different test cases in this paper series. One is adaptive navigation (Paper II) and another one in resilient motor control (Paper III). A fourth paper in these series presents a case in Business Intelligence to show the generality of the $SCAI$ framework across widely different environments.

\subsection{Discussion}
This paper definitely emphasizes the importance of the closed interaction loop between the agent and its environment in a dynamical way. 
Many researchers have clarified that intelligence resides in the circular relationship between the mind of an individual organism, its body, and the environment (Beer, 1995; Chiel \& Beer, 1997; Nolfi \& Floreano, 2000; Nolfi, Ikegami, \& Tani, 2008; Pfeifer \& Scheier, 1999; Tani, 1996; Varela, Thompson, \& Rosch, 1991). 
Also, an increasing number of research directions emphasize the embodied agent framework (Brooks \& Stein, 1994; Husbands, 2009; Lund, 2014). 
In this regard, different cognitive theories also emphasize the dynamic and interactive role of the brain and the environment  (Barandian, \& Moreno, 2006; Clark, 1997; Clark, \& Grush, 1999; 
Di Paolo, Barandiaran, Beaton \& Buhrmann, 2013;
Nolfi \& Tani, 1999; van Duijn, Keijzer, \& Franken, 2006; Vakarelov, 2011).

This paper has also emphasized the individual's self-construction of reality, 
yet the construction of reality has important social implications in higher species (Arbib \& Hesse, 1986; Butz, 2008; Piaget, 1954).
The agent lives in a society and the interaction with other agents plays an important role in the agent dynamics. 
Future work involves the extension to include social interactions. 
In this regard, an increasing body of research deals with the formalization of social interactions. 
For instance, Di Paolo \& De Jaegher (2012) introduce the interactive brain hypothesis in order to map the spectrum of possible relations between social interaction and neuronal processes. This hypothesis, among other things, states that interactive experience and skills play enabling roles in both the development and current function of social brain mechanisms, even in cases where social understanding happens in the absence of immediate interaction. 
In this regard, mirror neurons have been related to what to expect from other agents by direct observation, that is, they are related to the expectation of other agents' actions.
Hence, internal forward models have also been related to mirror neuron systems (Orzop, Kawato \& Arbib, 2013; Orzop, Wolpert \& Kawato, 2005; Tani, Ito, \& Sugita, 2004) and predictive coding (Grossberg, 2009; Kilner,  Friston \& Frith, 2007). 
Also, in the field of autonomous robots,  Steels and Spranger (2008) explain how autonomous robots can construct a body image of themselves and how this internal mental model, in turn, can help in motor control, planning and  in recognizing actions performed by other agents. 
Bostrom (2014) poses the implications of a futuristic superintelligence and the possible threats for the society if certain principles can not be guaranteed. Thus, it will be a very profound challenge to develop frameworks such as $SCAI$ for general artificial intelligence while making sure that certain ethical principles are guaranteed.

\section{Acknowledgments:} The author wishes to acknowledge Michael A. Arbib and Christoph von der Malsburg for very enlightening discussions on artificial intelligence, neural computation and schema-based learning during the author's Ph.D. residence in the Neural, Informational and Behavioral Sciences (NIBS) program at the University of Southern California (Los Angeles).

\section{References}

Adami, C. (2006). What do robots dream of? Science, 314, 1093-1094. 
\newline
\newline
Atkeson, C. G. (1989). Learning arm kinematics and dynamics. 
Annual Review of Neurosci., 12, 157-183. 
\\
\\
Albus, J. S. (2001). Engineering of mind: An introduction to the science of intelligent systems. Wiley.
\\
\\
Arbib, M. A. (1972). The Metaphorical Brain: An Introduction to Cybernetics as Artificial Intelligence and Brain Theory. 
New York: Wiley-Interscience.
\\
\\
Arbib, M. A. (1976). Artificial Intelligence: Cooperative Computation and Man-Machine Symbiosis.
 IEEE Transactions on Computers, 25(12), 1346-1352. 
\\
\\
Arbib. M. A. (1995). Schema Theory. In: Handbook of Brain Theory and Neural Networks. (ed. Michael A. Arbib). MIT Press. 
\\
\\
Arbib, M. A. \& Hesse, M. B. (1986). The construction of reality.  Cambridge University Press. 
\\
\\
Arbib, M. A., Lieblich, I. (1977). Motivational learning of spatial behavior. 
In: Systems Neuroscience, (ed. J. Metzler) pp. 119-165. New York: Academic Press.
\\
\\
Bach, J. (2009). Principles of synthetic intelligence PSI: an architecture of motivated cognition, volume 4. Oxford University Press.
\\
\\
Bartlett, F.C. (1932). Remembering, Cambridge University Press.
\\
\\
Beer, R. (1995). A dynamical systems perspective on agent-environment interaction. 
Artificial Intelligence, 72(1), 173-215.
\\
\\
Bongard, J. Zykov, V. \& Lipson, H. (2006). Resilient machines through continuous self-modeling.
 Science, 314, 1118-1121. 
\\
\\
Bostrom, N. 2014. Superintelligence: Paths, Dangers, Strategies. Oxford University Press.
\\
\\
Barandiaran, X. E., \& Moreno, A. (2006). On What Makes Certain Dynamical Systems Cognitive: 
A Minimally Cognitive Organization Program. Adaptive Behavior, 14(2), 171–185.
\\
\\
Brooks, R., \& Stein, L. A. (1994). Building brains for bodies. Autonomous Robots, 1, 7-25. 
\\
\\
Butz, M. V. (2008). How and why the brain lays the foundations for a conscious self.
Constructivist Foundations, 4, 1-42. 
\\
\\
Butz, M. V. (2010). Curiosity in learning sensorimotor maps. In: J. Haack, H. Wiese, A. Abraham, \& C. Chiarcos (Eds.). KogWis 2010 - 10. Tagung der Gesellschaft für Kognitionswissenschaft (p.92). Potsdam Cognitive Science Series 2. 
\\
\\
Butz, M. V., Herbort, O., \& Pezzulo, G. (2008). Anticipatory, goal-directed behavior. In G. Pezzulo, M. V.  Butz, C. Castelfranchi,  \& R. Falcone (Eds.) The Challenge of Anticipation: A Unifying Framework for the Analysis and Design of Artificial Cognitive Systems, LNAI 5225 (pp. 85-114).  Berlin Heidelberg: Springer
\\
\\
Butz, M. V.  \& Hoffman, J.  (2002). Anticipations control behavior: Animal behavior in an anticipatory learning classifier system. Adaptive Behavior, 10(2), 75-96.
\\
\\
Castelfranchi, C. (2005). Mind as an anticipatory device: For a theory of expectations. In: M. De
Gregorio, V. Di Maio, M. Frucci, C. Musio (eds.), Proceedings of Brain, Vision, and Artificial
Intelligence. Berlin: Springer, pp. 258-276.
\\
\\
Cagliori, D., Tommasino, P., Sperati, V., \& Baldassarre, G. (2014). Modular and hierarchical brain organization to understand assimilation, accomodation and their relation to autism in reaching tasks: a developmental robotics hypothesis. Adaptive Behavior,  22(5), 304-329. 
\\
\\
Chersi, F., Donnarumman, F. \& Pezzulo, G. (2013). Mental imagery in the navigation domain: a computational model of sensory-motor simulation mechanisms. Adaptive Behavior 21(4), 251-262. 
\\
\\
Chiel, H. \& Beer, R. (1997). The brain has a body: Adaptive behavior emerges from interactions of nervous system, body and environment. Trends in Neurosciences, 20, 553-557.
\\
\\
Clark, A. (1997). Being there: Putting brain, body and world together again. Cambridge, MA: MIT Press.
\\
\\
Clark, A. (2013). Whatever next? Predictive brains, situated agents, and the future of cognitive science. Behavioral and Brain Sciences, 36, 181-204.
\\
\\
Clark, A., \& Grush, R. (1999). Towards a Cognitive Robotics. Adaptive Behavior, 7(1), 5-16.
\\
\\
Corbacho, F. (1997). Schema-based Learning: Towards a Theory of 
Organization for Adaptive Autonomous Agents. Ph.D. Thesis. University of Southern  California, Los Angeles. 
\\
\\
Corbacho, F. (1998). Schema-based Learning. Artificial Intelligence, 101(1-2), 337-339.
\\
\\
Corbacho, F. (2016). Towards the Self-constructive Brain: emergence of adaptive behavior.  
http://arxiv.org/ abs/1608.02229.
\\
\\
Corbacho, F. \& Arbib, M. A. (1995). Learning to Detour. Adaptive Behavior, 3(4), 419-468.
\\
\\
Corbacho, F. \& Arbib, M. A. (1996). Schema-based Learmng for Adaptive Autonomous Agents. 
Proceedings of the First International Conference on Autonomous Agents. Los Angeles, CA. 
\\
\\
Corbacho, F. \& Arbib, M. A. (1997a). Towards a Coherence Theory of the Brain and Adaptive systems. Proceedings of the First International Conference on Vision, Recognition, and Action. (ed. Stephen Grossberg). Boston, MA.
\\
\\
Corbacho, F. \& Arbib, M. A. (1997b). Schema-based Learning: Biologically Inspired
Principles of Dynamic Organization. 
Lecture Notes in Computer Science 1240. Springer Verlag: Berlin.
\\
\\
Corbacho, F. J.; Nishikawa, K. C.; Weerasuriya, A., Liaw, J.S.; Arbib, M. A., (2005).
Schema-based learning of adaptable and flexible prey-catching in anurans. II. Learning after lesioning. 
Biological Cybernetics, 93(6), 410-425.
\\
\\
Craik, K. (1943). The Nature of Explanation. Cambridge University Press. 
\\
\\
Cruse, H., \& Steinkuehler, U. (1993). Solution of the direct and inverse kinematic problems by a commom algorithm based on the main multiple computations. Biological Cybernetics, 69, 345-351. 
\\
\\
Declerck, G. (2013). Why motor simulation can not explain affordance perception. Adaptive Behavior Journal, 21(4),286-298
\\
\\
Desmurget, M., \& Grafton, S. (2000). Forward modeling allows feedback control for fast reaching movements. Trends in Cognitive Sciences, V. 4(11), 
\\
\\
Di Nuovo, A. G., Marocco, D., Di Nuovo, S., \& Cangelosi, A. (2013). Autonomous learning in humanoid robotics through mental imagery. Neural Networks, 41, 147-155. 
\\
\\
Di Paolo, E. A., Barandiaran, X. E., Beaton, M. \& Buhrmann, T. (2013). Learning to perceive in the sensorimotor approach: Piaget´s theory of equilibration interpreted dynamically. Frontiers in Human Neuroscience, 8(551), 1-16. 
\\
\\
Di Paolo, E. A. \& De Jaegher, H. (2012). The interactive brain hypothesis. Frontiers in Human Neuroscience, 6(163), 1-16.
\\
\\
Drescher, G. L. (1991). Made-up minds: A constructivist approach to artificial intelligence. MIT Press.
\\
\\
 Driskell, J. E., Copper, C., \& Moran, A. (1994). Does mental practice enhance performance? Journal of Applied Psychology, 79(4), 481-492.
\\
\\
Droulesz, J., \& Berthoz, A. (1991). A neural network model of sensonitopic maps with predictive short-term properties. Proceedings of the National Academy of Sciences, U.S.A., 88, 9653-9657. 
\\
\\
Duhamel, J.R., Colby, C. L., \& Goldberg, M. E. (1992) The updating of the representation of visual space in parietal cortex by intended eye movements. Science, 255, 90-92
\\
\\
Flanagan, J., \& Johansson, R. (2003). Action plans used in action observation. Nature, 424, 769-771. 
\\
\\
 Flanagan, J. R., Vetter, P., Johansson, R. S. \& Wolpert, D. M. (2003). Prediction precedes control in motor learning. Curr. Biol. 13, 146-150. 
\\
\\
 Flanagan, J. R., \& Wing, A. M. (1997). The role of internal models in motion planning and control: 
evidence from grip force adjustments during movements of hand-held loads. J. Neurosci. 17, 1519-1528. 
\\
\\
Friston, K. J. (2010). The free-energy principle: a unified brain theory. Nature Neuroscience, 11, 127-138. 
\\
\\
 Garcia, C. E., Prett, D. M., \& Morari, M. (1989). Model Predictive Control: Theory and Practice- a survey. Automatica, 25, 335-348.
\\
\\
Gentili, R., Han, C. E., Schweighofer, N., \& Papaxanthis, C. (2010). Motor learning without doing: Trial-by-trial improvement in motor performance during mental training. J. Neurophysiol., 104, 774-783. 
\\
\\
 Gilbert, D.  T. \& Wilson, T. D. (2007). Prospection: experiencing the future. Science, 317, 1351-1354.
\\
\\
Goertzel, B., and Pennachin, C. 2007. Artificial General Intelligence. Springer.
\\
\\
Goertzel, B.; Lian, R.; Arel, I.; de Garis, H.; and Chen, S. 2010. A world survey of artificial brain projects, Part II: Biologically inspired cognitive architectures. Neurocomputing 74(1):30-49.
\\
\\
Goertzel, B. 2010. Toward a formal characterization of real-world general intelligence. In Proceedings of the Third Conference on Artificial General Intelligence, 19-24.
\\
\\
Goertzel, B. (2014). Artificial General Intelligence: Concept, State of the Art, and Future Prospects. Journal of Artificial General Intelligence 5(1), 1-46. 
\\
\\
Gregory, R. L. (1967). Will seeing machines have illusions? In Machine Intelligence 1 (N. L. Collins and D. Michie eds.). Oliver \& Boyd. 
\\
\\
Grossberg, S. (2009). Cortical and subcortical predictive dynamics and learning during perception, cognition, emotion and action. Phil. Trans. R. Soc. B., 364, 1223-1234. 
\\
\\
 Guazzelli, A, Corbacho, F., Bota, M. \& Arbib, M. A. (1998). An implementation of the Taxon-Affordance System for Spatial Navigation.  Adaptive Behavior, 
6(4), 435-471.  
\\
\\
 Haruno, M., Wolpert, D. M. \& Kawato, M. (2001). MOSAIC model for sensorimotor learning and control. Neural Computation, 13, 2201-2220. 
\\
\\
Hassabis, D. \& Maguirre, E. A. (2009). The construction system of the brain. Philosophical Transactions of the Royal Society B. Theme Issue ‘Predictions in the brain: using our past to prepare for the future’ compiled by Moshe Bar.
\\
\\
Hawkins, J.  \& Blakeslee, S. (2004). On intelligence. Times Books: New York.
\\
\\
 Hesslow, G. (2002). Conscious thought as simulation of behavior and perception. Trends in Cognitive Sciences, 6, 242-247.
\\
\\
Hesslow, G. (2012). The current status of the simulation theory of cognition. Brain Research, 1428, 71-79. 
\\
\\
 Hinton, G. E., Dayan, P., Frey, B. J. \& Neal, R. (1995). 
The wake-sleep algorithm for unsupervised Neural Networks. Science, 268, 1158-1161. 
\\
\\
Hinton, G. E., Krizhevsky, A. \& Wang, S. (2011). Transforming Auto-encoders. ICANN-11: International Conference on Artificial Neural Networks, Helsinki. 
\\
\\
 Husbands, P. (2009).  Never mind the Iguana, What about the Tortoise? Models in Adaptive Behavior. Adaptive Behavior, 17(4), 320-324. 
\\
\\
Jacobs, R. A, Jordan, M. I., Nowlan, S. J., \& Hinton, G. E. (1991). Adaptive mixtures of local
experts. Neural Computation, 3, 79-87.
\\
\\
Jeannerod, M. (2001). Neural simulation of action: a unifying mechanism for motor cognition. Neuroimage, 14, 103-109. 
\\
\\
Johnson-Laird, X, Y.  (1983). Mental models: Towards a cognitive science of language, inference, and consciousness. Cambridge. Cambridge University Press and Harvard University Press. 
\\
\\
Jordan, M.I. (1983). Mental practice. Unpublished dissertation proposal, Center for Human Information Processing, University of California, San Diego.
\\
\\
Jordan, M. I. \& Rumelhart, D. (1992). Forward models:
Supervised Learning with a Distal Teacher. Cognitive Science, 16, 307-354.
\\
\\
Jordan, M. I., \& Jacobs, R. A. (1992). Hierarchies of adaptive experts. In J. Moody, S. Hanson,
\& R. Lippmann (Eds.), Advances in Neural Information Processing Systems 4. San Mateo,
CA: Morgan Kaufmann. pp. 985-993.
\\
\\
Kalman, R. E. (1960). A new approach to linear filtering and
prediction theory. Trans. ASME J. Basic Eng., 82, 35-45.
\\
\\
Kawato, M. (1999) Internal models for motor control and trajectory planning. Curr. Opin. Neurobiol. 9, 718-727
\\
\\
Kawato, M. (1990a). Feedback-error-learning neural network for supervised
learning. In R. Eckmiller (Ed.), Advanced neural computers (pp. 365-372). Amsterdam: North-Holland.
\\
\\
Kawato, M. (1990b). Computational schemes and neural network models for formation and control of multijoint arm trajectory. In W.T. Miller, III, R.S. Sutton, \& P.J. Werbos
(Eds.), Neural networks for control. Cambridge, MA: MIT Press.
\\
\\
Kawato, M., Furukawa, K., \& Suzuki, R. (1987). A hierarchical neural network model for control and learning of voluntary movement. Biological Cybernetics, 57, 169-185.
\\
\\
Kemp, C. \& Tenenbaum, J. B. (2008). The discovery of structural form. Proc Nat Acad. Sci. 105(31): 10687-10692.
\\
\\
 Kilner, J. M., Friston, K. J., \& Frith, C. D. (2007). Predictive coding: an account of the mirror neuron system. Cognitive Processing 8, 159-166.
\\
\\
 Kossyln, S. M., Gani, G. \& Thompson, W. L. (2001). Neural foundations of imagery. Nature Reviews Neurosci. 2(9), 635-642.
\\
\\
Kurzweil, R. (2005). The Singularity is Near, New York: Viking Books, ISBN 978-0-670-03384-3.
\\
\\
Kurzweil, R. (2012). How to Create a Mind: The Secret of Human Thought Revealed, New York: Viking Books, ISBN 978-0-670-02529-9.
\\
\\
Lallee, S. \&  Dominey, P. F. (2013). Multi-modal convergence maps: from body schema and self-representation to mental imagery. Adaptive Behavior 21(4), 274-285.
\\
\\
 LeCun, Y., Bengio, Y., \& Hinton, G. (2015). Deep Learning. Nature, 521: 436-444. 
\\
\\
Legg, S; Hutter, M (2007). A Collection of Definitions of Intelligence (Technical report). IDSIA. arXiv:0706.3639 
\\
\\
 Lieblich, I. \& Arbib, M. A. (1982). Multiple representations of space underlying behavior. Brain Behav. Sci., 5: 627-659.
\\
\\
Lungarella, M., Metta, G., Pfeifer, R. \& Sandini, G. (2003) Developmental robotics: a survey. Connect. Sci. 15(4), 151–190.
\\
\\
Luger, G. \& Stubblefield, W. (2004). Artificial Intelligence: Structures and Strategies for Complex Problem Solving. 5th Edition, The Benjamin Cummings Publishing Company. 
\\
\\
 Lyons D. \& Arbib, M. A. (1989). A formal model of Distributed computation for Scjhema-based robot control. IEEE
J. Robotics and Automation, 5:280-293.
\\
\\
Mehta, B. \& Schaal, S. (2002).  Forward models in visuomotor control. Journal of Neurophysiology 88, 942-953.
\\
\\
Merfeld, D. Zupan, L. \& Peterka, R. J. (1999). Humans use
Internal Models to Estimate gravity and linear acceleration. Nature,
398:615.
\\
\\
Miall, R.C. \& Wolpert, D. M. (1996). Forward Models for Physiological Motor Control. Neural Networks. 9(8): 1265-1279. 
\\
\\
Mohan, V., Sandini, G. \& Morasso, P. (2014). A Neural Framework for Organization and Flexible Utilization of Episodic Memory in Cumulatively Learning Baby Humanoids. Neural Computation 26, 2692-2734. 
\\
\\
 Mussa-Ivaldi F. A., Bizzi E. (2000). Motor learning through the combination of primitives. Philos Trans R Soc Lond B Biol Sci 355(1404):1755-69. 
\\
\\
 Mussa-Ivaldi, F. A. (1999). Modular features of motor control and learning. Current Opinion in Neurobiology, vol. 9, pp. 713-717.
\\
\\
Neisser, U. (1976).  Cognition and Reality: Principles and Implications of Cognitive Psychology, W.H. Freeman.
\\
\\
Nilsson, N. (1998). Artificial Intelligence: A New Synthesis. Morgan Kaufmann Publishers, San Francisco, CA,
\\
\\
Nolfi, S., \& Floreano, D. (2000). Evolutionary robotics: The biology, intelligence, and technology of self-organizing machines. Cambridge, MA: MIT Press. 
\\
\\
 Nolfi, S., Ikegami, T. \& Tani, J. (2008). Behavior and Mind as a Compex Adaptive system. Adaptive Behavior, 16(2), 101-103. 
\\
\\
 Nolfi, S., \& Tani, J. (1999). Extracting regularities in space and time through a cascade of prediction networks: The case of a mobile robot navigating in a structured environment. Connection Science, 11, 125-148. 
\\
\\
 Oztop, E., Kawato, M., \& Arbib, M. A. (2013). Mirror neurons: Functions, mechanisms and models. Neurosci. Letters 540, 43-55.
\\
\\
 Oztop, E., Wolpert, D. \& Kawato, M., (2005). Mental state inference using visual control parameters. Cognitive Brain Research 22, 129-151. 
\\
\\
 Pezzulo, G. (2008). Coordinating with the future: The Anticipatory Nature of Representation. Minds and Machines, 18, 179-225. 
\\
\\
 Pezzulo, G., Butz, M. V., Sigaud, O., \& Baldassarra, G.  (2009). Anticipatory behavior in adaptive learning systems. LNCS 5499.
\\
\\
Pezzulo, G., Candidi, M., Dindo, H., \& Barca, L. (2013). Action simulation in the human brain: Twelve questions. New Ideas in Psychology, 1-21.
\\
\\
 Pfeifer, R. \& Scheier, C. (1999). Understanding intelligence. Cambridge, MA: MIT Press.
\\
\\
Piaget, J. (1954). The construction of reality in the child. Ballantine. New York. 
\\
\\
Piaget, J. (1971). Biology and Knowledge, Edinburgh University Press, Edinburgh.
\\
\\
Poole, D., Makworth, A. \& Goebel, R.  (1998). Computational Intelligence: A Logical Approach. Oxford University Press. 
\\
\\
 Raos, V., Evangeliou, M.N., \& Savaki, H. E. (2007). Mental simulation of action in the service of action perception. The Journal of Neuroscience 27, 12675-12683. 
\\
\\
 Redon, C. et al. (1991) Proprioceptive control of goal directed movements in man studied by means of vibratory muscle tendon stimulation. J. Mot. Behav. 23, 101–108. 
\\
\\
Rosen, R. (1985). Anticipatory systems. Oxford: Pergamon Press. 
\\
\\
Rosenbloom, P. S. (2013). The Sigma cognitive architecture and system. AISB Quarterly, 136, 4-13. 
\\
\\
Russell, S. \& Norvig, P. (2010). Artificial Intelligence: A Modern Approach. Prentice Hall (3rd edition). 
\\
\\
Sabour, S., Frosst, N. \& Hinton, G. E. (2017).  Dynamic Routing between Capsules. 
31st Conference on Neural Information Processing Systems (NIPS 2017), Long Beach, CA, USA.
\\
\\
Salakhutdinov, R., Tenenbaum, J. B. \& Torralba, A.  (2013). Learning with Hierarchical-Deep Models.
IEEE Transactions on Pattern Analysis and Machine Intelligence, 35(8), 1958-1971.
\\
\\
Sanchez-Montanes, M., \& Corbacho, F. (2004). A new Information Processing Measure for Adaptive Complex Systems. 
IEEE Trans. Neural Netw., 15(4), 917-927. 
\\
\\
Sanchez-Montanes, M., \& Corbacho, F. (2019). Structural Learning: a general framework and a specific implementation. 
(in preparation). 
\\
\\
Shadmehr, R., \& Mussa-Ivaldi, F. (1994). Adaptive representation of dynamics during learning of a motor task. Journal of Neuroscience, 14, 3208-3224. 
\\
\\
Shadmehr, R., Smith, M. A., \& Krakauer, J. W. (2010). Error correction, sensory prediction, and adaptation in motor control. Annual Review of Neuroscience, 33, 89-108. 
\\
\\
Shen, W. M. (1994). Autonomous Learning from the Environment. W H Freeman \& Co.
\\
\\
Sigaud, O. Butz, M. V. Pezzulo, G. \& Herbort, O. (2013). The anticipatory construction of reality as a central concern for psychology and robotics. New Ideas in Psychology, 31(3), 217-220. 
\\
\\
Smith-Bize, S. C. (2017). Self-Organisation of Internal Models in Autonomous Robots. Doctoral Dissertation. The University of Edinburgh. 
\\
\\
 Steels, L. \& Spranger, M. (2008). The robot in the mirror. Connection Science, 20(4), 337-358. 
\\
\\
 Suri, R. E., \& Schultz, W. (2001).Temporal Difference Model
 Reproduces Anticipatory Neural Activity. Neural Computation, 13(4), 841-862.
\\
\\
 Sutton, R.S. (1988). Learning to predict by the methods of temporal differences. Machine Learning, 3, 9-44.
\\
\\
 Sutton, R.S. (1990). Integrated architectures for learning, planning, and reacting based on approximating dynamic programming. Proceedings of the Seventh International Conference
on Machine Learning pp. 216-224.
\\
\\
 Sutton, R. S., \& Barto, A. G. (1981). An adaptive network
that constructs and uses an internal model of its world. Cognition and
Brain Theory, 4(3): 217-246.
\\
\\
 Sutton, R. S., \& Barto, A. G. (1998). Reinforcement learning: An introduction. MIT Press. 
\\
\\
 Svensson, H., Thill, S., \& Ziemke, T. (2013). Dreaming of electric sheep? Exploring the functions of dream-like mechanisms in the development of mental imagery simulations. 
Adaptive Behavior 21(4), 222-238.
\\
\\
 Tani, J. (1996). Model-based Learning for mobile robot navigation from the dynamical systems perspective. IEEE Transactions on Systems, Man, and Cybernetics B, 26(3), 421-436.
\\
\\
Tani, J., Ito, M., \& Sugita, Y. (2004). Self-organization of distributedly represented multiple behavior schemata in a mirror system: reviews of robot experiments using RNNPB, Neural Networks, 17, 1273-1289. 
\\
\\
Taylor, M. E. Khulmann, G. \& Stone, P.  (2008). Autonomous Transfer for Reinforcement Learning. 
Proc. of 7th Int. Conf. on Autonomous
Agents and Multiagent Systems (AAMAS 2008).
\\
\\
Tenenbaum, J. B., Kemp, C, Griffiths, T. L. \& Goodman, N.   (2011). How to Grow a Mind: Statistics, Structure, and Abstraction. Science, 331, 1279-1285. 
\\
\\
Torralba, A. (2003). Contextual Priming for Object Detection. International Journal of Computer Vision, 53(2), 169-191.
\\
\\
Vakarelov, O. (2011). The cognitive agent: Overcoming informational limits. Adaptive Behavior, 19(2), 83-100. 
\\
\\
 van Duijn, M., Keijzer, F. \& Franken, D. (2006). Principles of Minimal Cognition: Casting Cognition as Sensorimotor Coordination. Adaptive Behavior, 14(2), 157-170.
\\
\\
 Varela, F., Thompson, E., \& Rosch, E. (1991). The embodied mind. Cambridge, MA: MIT Press. 
\\
\\
 Wada, Y., \& Kawato, M. (1993). A neural network model for arm trajectory formation using forward and inverse dynamics models. Neural Networks, 6, 919-932.
\\
\\
Weitzenfeld, A., Arkin, R.C., Cervantes-Perez, F., Olivares, R., \&  Corbacho, F. (1998).
A Neural Schema Architecture for Autonomous Robots.
Proc. International Symposium on Robotics and Automation. Mexico.
\\
\\
Weng, J., McClelland, J., Pentland, A., Sporns, O., Stockman, I., Sur, M. \& Thelen, E. (2001). Autonomous mental development by robots and animals. Science, 291, 599-600.
\\
\\
Wolpert, D. M., Ghahramani, Z., \& Jordan, M. I. (1995). An
Internal model for sensorimotor integration. Science, 269, 1880-1882.
\\
\\
Wolpert, D. M., \& Kawato, M. (1998). Multiple paired forward and inverse models for motor control. Neural Networks 11, 1317-1329.


\end{singlespacing}

\end{document}